\documentclass{article}

\PassOptionsToPackage{numbers, compress}{natbib}

\usepackage{microtype}
\usepackage{graphicx}
\usepackage{subcaption}
\usepackage{booktabs} 
\usepackage{hyperref}
\usepackage{wrapfig}
\usepackage{stfloats}       
\usepackage{multicol}
\usepackage{multirow}
\usepackage{comment}
\usepackage{enumitem}
\usepackage{bbm}
\usepackage{siunitx}
\usepackage{algorithm}
\usepackage{algpseudocode}
\usepackage{xcolor}
\usepackage[most]{tcolorbox}
\usepackage{mathtools}
\usepackage{longtable}
\usepackage{wrapfig}
\tcbuselibrary{breakable}
\algrenewcommand\algorithmicindent{1.2em}
\algrenewcommand\algorithmiccomment[1]{\hfill$\triangleright$ #1}

\usepackage[preprint]{neurips_2026}

\usepackage[utf8]{inputenc} 
\usepackage[T1]{fontenc}    
\usepackage{hyperref}       
\usepackage{url}            
\usepackage{booktabs}       
\usepackage{amsfonts}       
\usepackage{nicefrac}       
\usepackage{microtype}      
\usepackage{xcolor}         

\title{Abstraction for Offline\\Goal-Conditioned Reinforcement Learning}

\author{%
  Clarisse Wibault \\
  FLAIR, MLRG\\
  University of Oxford\\
  \footnotesize{clarisse.wibault@magd.ox.ac.uk}\\
  \And
  Alexander Goldie \\
  FLAIR \\
  University of Oxford \\
  \And
  Antonio Villares \\
  FLAIR \\
  University of Oxford \\
  \AND
  Maike Osborne\\
  MLRG \\
  University of Oxford \\
  \And
  Jakob Foerster \\
  FLAIR \\
  University of Oxford \\
}

\begin{document}

\maketitle

\begin{abstract}
Markov Decision Processes (MDPs) often exhibit significant redundancy due to symmetries and shared structure across state-goal pairs in real-world Goal-Conditioned Reinforcement Learning (GCRL). While hierarchical policies have been motivated for horizon reduction via \emph{temporal} abstraction in offline GCRL, we demonstrate that hierarchy also enables \emph{absolute} abstraction. By introducing \emph{relativised} options as well as \emph{distinct representations} for different levels of the hierarchy, we demonstrate how an agent can reuse experience across similar contexts of the state-space. Based on this framework, we introduce two simple algorithms for learning relativised options and abstracting from the absolute frame of reference. Our experiments show that such inductive biases significantly improve performance in offline GCRL. 
\begin{figure}[H]
    \centering
    \includegraphics[width=0.8\linewidth]{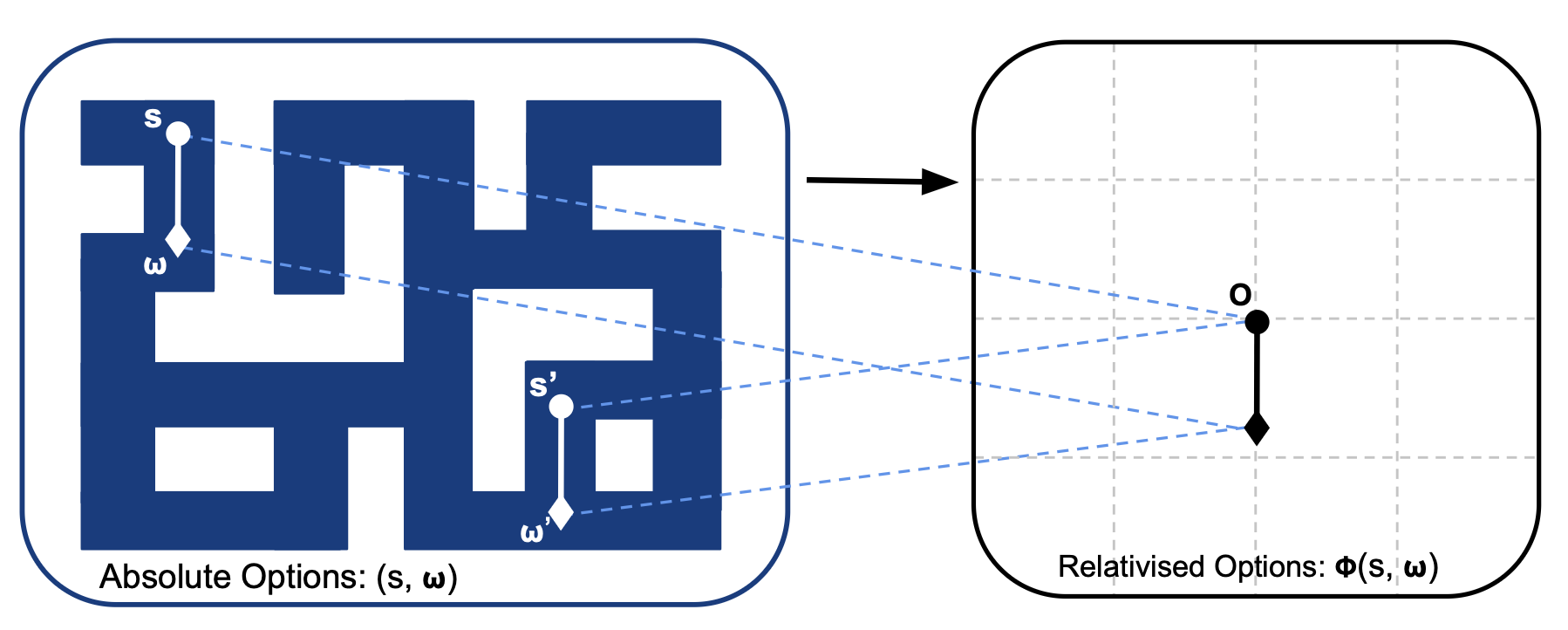}
    \caption{\textbf{Abstractive RL (ARL).} By learning relativised options, ARL enables the reuse of experience across similar contexts of the state-space.}
    \label{fig: arl}
\end{figure}
\end{abstract}

\section{Introduction}
\label{sec: intro}
Offline Goal-Conditioned Reinforcement Learning (GCRL) \citep{kaelbling_learning_1993, schaul2015universal, levine_offline_2020, levine_understanding_2021, park_ogbench_2025} provides a principled framework for training a general-purpose agent to solve complex long-horizon tasks from static datasets. However, in practice, existing methods have struggled to learn effective policies from offline data, partly due to imperfect dataset coverage of the state-action space \cite{levine_offline_2020, prudencio_survey_2024}. Furthermore, since common offline RL algorithms \citep{kostrikov_offline_2021, peng_advantage-weighted_2019, tarasov_revisiting_2023} regularise actions towards those close to the dataset distribution to mitigate issues such as distribution shift \citep{levine_offline_2020, prudencio_survey_2024}, an agent may fail to recover an optimal policy if a dataset only contains low-quality actions in certain regions of the state space. This is a key challenge in offline RL \cite{dulac-arnold_challenges_2019} and leads to difficulties in value estimation, policy extraction, and policy generalisation \citep{park_is_2024}.

Recent work suggests that horizon reduction is essential for scaling offline RL \citep{park_horizon_2025}, motivating the use of hierarchical policies, or options \citep{sutton_between_1999}, for temporal abstraction \citep{vezhnevets2017feudal}. We extend this perspective by arguing that hierarchy offers an additional advantage: \emph{absolute} abstraction. By using \emph{absolute} abstraction, which we define as using \emph{relativised} options and \emph{distinct representations} at different levels of the hierarchy, an agent abstracts away from the absolute frame of reference and can reuse experience across similar contexts of the state-space.

To illustrate this point, consider an agent undertaking a locomotion task. While the dataset may only contain a limited number of suboptimal demonstrations of the full task, many of the constituent relativised options (subtasks such as simply moving forward or navigating a corner) might be well represented across demonstrations with different goals. By defining options relative to the agent's local context (e.g. \emph{navigate to the corner directly ahead} rather than \emph{navigate to corner A}) and decoupling low-level actions (i.e motor-actuation) from redundant high-level information, the agent can leverage many more subtask examples to learn a policy.

In principle, relativised options \citep{ravindran_model_2004} exploit redundancy and symmetry in MDPs to allow behaviours to generalise across states. However, in practice, implementations remain limited to toy examples due to the inherent difficulty of identifying MDP homomorphisms or learning such relative representations. In this work, we introduce \emph{Abstractive Reinforcement Learning} (ARL), a general framework that learns relativised options via action similarity. Based on this, it defines high-level similarity and low-level similarity respectively as state-goal pairs inducing similar options and state-option pairs inducing similar immediate actions. 

We propose two simple algorithms that comply with the ARL framework: the first can be applied generally, simply using action similarity to implicitly learn relativised options; the second introduces a representational inductive bias for the low-level MDP by explicitly enforcing translational invariance, improving generalisation in certain high-dimensional manipulation tasks. Our experiments demonstrate that such relativised options and inductive biases result in better performance in high-dimensional offline GCRL.

\textbf{Contributions.} Concisely, this work addresses the following question: 
\begin{center} \textit{Can hierarchy enable more robust policy extraction in regions where the dataset suffers from low-quality transitions?} \end{center}
Our contributions are two-fold. We first motivate hierarchy in offline RL through abstraction from the absolute frame of reference. We show how relativised options and distinct representations at different levels of the hierarchy can enable data reuse, bounding the maximum error in expected return compared to a flat policy. Secondly, we introduce two simple algorithms that learn relativised options and abstract from the absolute frame of reference. These algorithms outperform both flat policies and hierarchical ones that are anchored in the absolute state-space in high-dimensional tasks, without introducing additional hyperparameters.

Explicitly, for an RL practitioner, our work demonstrates that: (i) options should be learned via \emph{action similarity} rather than value similarity i.e. jointly optimised with the low-level policy; (ii) we necessitate two value functions to decouple the high-level from the low-level decision process; and (iii) that imposing translation invariance on the low-level MDP can improve policy generalisation in high-dimensional manipulation tasks. For an RL researcher, our work opens up new avenues, such as methods to learn relativised options via action chunking \cite{black_real-time_2025, park_scalable_2025}, or incorporating more flexible inductive biases using ideas from Geometric Deep Learning (GDL) \citep{bronstein2021geometric, tangri2025equivariant}.  

\section{Preliminaries}
\label{sec: prelims}
In Offline \citep{levine_offline_2020, lange_2012_batchrl} Goal-Conditioned Reinforcement Learning (GCRL) \cite{kaelbling_learning_1993} an agent seeks to learn a universal policy \citep{schaul2015universal} from a fixed dataset, enabling it to reach arbitrary goal states in the smallest number of timesteps \citep{park_ogbench_2025}.

\subsection{Problem Setting}
\label{sec: prelims_problem_setting}
We consider a standard Markov Decision Process (MDP) \citep{sutton_2018_reinforcement} defined by the tuple $\mathcal{M} :=(p_{s_0}, \mathcal{S}, \mathcal{G}, \mathcal{A},  \mathcal{T}, \beta_g, \gamma)$, where $p_{s_0}$ is the initial state distribution, $\mathcal{S}$, $\mathcal{G}$ and $\mathcal{A}$ respectively denote the state, goal and action space, $\mathcal{T}$ is the transition function, $\beta_g$ is a goal-conditioned pseudo-termination function, and $\gamma<1$ is the discount factor. At the beginning of the episode, a state $s_0$ is sampled from $p_{s_0}$. A goal state $g$ is uniformly sampled from the goal space $\mathcal{G}$, and is fixed for the entire episode. The goal space may be defined over all or a subset of the state dimensions. At each timestep $t \ge 0$, an agent takes an action $a_t$ conditioned on its current state $s_t$ and goal state $g$, and transitions to a new state $s_{t+1} = \mathcal{T}(\cdot \mid s_t, a_t)$. Episodes terminate according to the goal-conditioned pseudo-termination function \cite{white_scaling_2012} $\beta_g : \mathcal{S} \to \{0,1\}$, where $\beta_g(s) = 1$ if and only if the goal has been reached. Following \citet{andrychowicz_hindsight_2018}, we focus on the problem of sparse and binary rewards, which is motivated in robotics, for example. The agent receives a reward of $-1$ on all steps, and a reward of $0$ upon reaching the goal $r_t = - \mathbbm{1} \{\beta_g(s_t) = 0\}$. The aim of the agent is to learn a universal policy \citep{schaul2015universal} conditioned on its state and the goal $\pi: \mathcal{S} \times \mathcal{G} \to \Delta(\mathcal{A})$ that maximises the sum of discounted returns $\mathbb{E}_\pi\left[\sum_{t=0}^\infty \gamma^tr_{t}\right]$ from a fixed dataset $\mathcal{D}$. The dataset contains $N$ state-action trajectories of length $H$, collected using an arbitrary policy in a task-agnostic manner. Following \citet{park_hiql_2024}, we adopt the fully observed deterministic MDP framework for simplicity, though we additionally evaluate in a stochastic environment.

\subsection{Offline Reinforcement Learning}
Offline RL is commonly formulated as a two-step procedure: first, value learning, and subsequently policy extraction from this value function. A major challenge in offline RL is extracting an optimal policy from suboptimal data. This difficulty is exacerbated by distribution shift \cite{levine_offline_2020, dulac-arnold_challenges_2019}, where the learned policy induces state-action pairs that are poorly supported by the dataset, leading to unreliable value estimates.

To address this issue, model-free algorithms typically incorporate some form of conservatism during value learning, regularise policy extraction towards the dataset distribution via behaviour-cloning, or combine both strategies. The value function might be learned with a distribution-constrained objective \citep{kumar_conservative_2020, kostrikov_offline_2021}, to minimise uncertainty \citep{an_uncertainty-based_2021} or encourage structured representations \citep{eysenbach_contrastive_2023}. Similarly for policy extraction, the agent is regularised by using an additional behaviour-cloning loss term \citep{fujimoto_minimalist_2021}, weighted behaviour-cloning \citep{peng_advantage-weighted_2019}, or rejection-sampling \citep{ghasemipour_expected_2007} of a behaviour-cloned policy. In all cases: 
\begin{center}
    \textit{Due to their underlying bias towards the dataset distribution, flat offline RL algorithms struggle in regions where the dataset suffers from low-quality transitions.}
\end{center}

\subsection{Options}
Options \citep{sutton_between_1999} provide a framework for decision making at different levels of temporal abstraction. In the original options framework, each option $\omega \sim \Omega$ represents temporally extended behaviour, executed until termination according to a learned or predefined termination condition \cite{bacon_option-critic_2016}. 
In contrast, hierarchical approaches in Offline GCRL typically resample options at every timestep \citep{park_hiql_2024, park_horizon_2025, baek_graph-assisted_2025}, effectively removing temporal commitment and avoiding the need to specify a termination function:
\begin{equation}
    \pi(a \mid s,g) = \pi_l(a \mid s, \omega) \quad \omega \sim \pi_h(\cdot \mid s,g),
    \notag
\end{equation}
where $\pi_h$ is the high-level policy that samples an option conditioned on the current state and goal, and $\pi_l$ is a low-level policy that samples an action conditioned on the same state and that sampled option. Such approaches still differ from a flat policy, since the low-level policy is conditioned on the option rather than the goal. Since local gradient information can be uninformative or misleading when optimising for distant goals, a key benefit of this hierarchical inductive bias is that it reduces the effective horizon for both the high-level and low-level MDP \citep{park_hiql_2024, park_horizon_2025}.

\section{Related Work}
\label{sec: related_work}
Extensive literature motivates hierarchy for temporal abstraction and policy horizon reduction \citep{sutton_between_1999, sutton_intra-option_nodate, ravindran_smdp_2003}, but apart from \citet{nachum_data-efficient_2018}, \citep{nachum_near-optimal_2019} and \citet{levy_learning_2019}, very little work explicitly motivates hierarchy via representation learning and data reuse. While original work used a set of hardcoded options \cite{sutton_between_1999}, since then options have been learned by identifying bottleneck states (e.g. \citep{mcgovern2001automatic, stolle2002learning}), or by jointly learning the options with the low-level policy \citep{bacon_option-critic_2016}. Most work defines options in the original state-space, but, for example, \citet{vezhnevets2017feudal} learn options in an embedding space. \citet{ravindran_model_2004} introduce the concept of relativised options, but non-hardcoded implementations of relativised options have remained scarce. We remark that options can be defined in a latent space, but still be anchored to the absolute frame of reference. All of these works study hierarchical RL in the online setting, while we focus on the offline setting. 

Offline GCRL naturally lends itself to hierarchical formulations, where high-level policies specify intermediate goals, and low-level policies learn intermediate goal-reaching behaviours \citep{park_ogbench_2025, park_is_2024, park_horizon_2025, park_hiql_2024, baek_graph-assisted_2025, li2022hierarchicalplanninggoalconditionedoffline}. In hierarchical offline GCRL, options have again generally taken the form of absolute options in the original state-space (e.g. \citep{park_horizon_2025,baek_graph-assisted_2025}). Our approach is most related to HIQL \citep{park_hiql_2024}, which learns latent options, but differs in two fundamental ways. First, unlike HIQL, which uses one value function, we use two distinct value functions, enabling different representations at different levels of the hierarchy (Section \ref{sec: implementation_2}). Second, rather than basing option embeddings on value similarity, we base them on action similarity (Section \ref{sec: implementation_1}). This changes the organisation of the latent space, since two state-waypoints might induce similar values but differ entirely in their low-level actions.    

The theory of state abstraction identifies conditions under which information can be compressed while maintaining policy optimality at different levels of abstraction \citep{li_towards_nodate}. Prior work in State Representation Learning (SRL) has explored homomorphisms, bisimulation metrics, and contrastive objectives to map together semantically similar states (e.g. \citep{ravindran_smdp_2003, ferns_metrics_2012, chen_learning_nodate}). However, these methods often require auxiliary loss terms that necessitate hyperparameter tuning and can lead to training instability. We refer to \citet{echchahed_survey_2025} for an overview. Furthermore, none of these methods explicitly address hierarchical goal-conditioned representation learning.

\section{Abstractive Reinforcement Learning (ARL)}
\label{sec: method}
In this section, we introduce Abstractive Reinforcement Learning (ARL), a framework for learning abstractions in offline hierarchical GCRL to improve robustness in regions where the dataset suffers from low-quality transitions. Based on this framework, we introduce two simple algorithms: the first learns relativised options via action similarity, while the second additionally imposes translational invariance on the low-level MDP. Together, these algorithms demonstrate how (i) relativised options and (ii) representational inductive biases can improve generalisation in offline GCRL.

\subsection{Objective}
\label{sec: objective}
In principle, our aim is to learn abstractions that enable reuse of experience across similar contexts. By learning options that group together state-waypoint pairs, which, under an optimal policy, induce similar action sequences, options $\omega \in \Omega$ are relative rather than anchored to an absolute frame of reference. This naturally induces two hierarchical notions of similarity: a high-level similarity, where similar state-goal pairs induce similar options, and a low-level similarity, where similar state-option pairs induce similar immediate actions. Consequently, we can define high-level embeddings $\phi_h(s,g)$ and low-level embeddings $\phi_l(s,\omega)$ that abstract away information irrelevant to their respective levels:
\begin{equation}
    \pi(a \mid s, g) = \pi_l( a \mid \phi_l(s, \omega)) \quad \omega \sim \pi_h(\cdot \mid \phi_h(s, g)).
\end{equation}
To allow the high-level and low-level decision processes to operate on different representations and at different temporal abstractions (i.e. different discount factors), such an approach necessitates two distinct value functions. In the following Motivation Box, we provide an intuition on how such abstractions can mitigate failures in regions where the dataset suffers from low-quality transitions by analysing the maximum error in offline RL for a finite-state, finite-action MDP.
\definecolor{clearsky}{RGB}{100,150,255}
\begin{tcolorbox}[
    breakable,
  colback=clearsky!10!white,
  colframe=clearsky!95!black,
  title= \textbf{Motivation Box:} Bounding the Maximum Error,
  boxrule=0.6pt,
  boxsep=2pt,
  left=4pt,
  right=4pt,
  top=4pt,
  bottom=4pt,
]
We consider learning an optimal policy in a finite-state, finite-action MDP from a fixed dataset $\mathcal{D}$ of size $N$. We refer to Appendix \ref{app: error} for a complete derivation and full definitions. We assume at most two possible next states for each state-action pair, which is realistic given our deterministic transition assumption (Section \ref{sec: prelims}). For a flat goal-conditioned policy, the Probably Approximately Correct (PAC) Learning error $\epsilon$ in expected return is given by \citep{kakade2003sample, munos_finite-time_nodate, lattimore_pac_2012}: 
\begin{equation}
    \epsilon \propto \sqrt{\frac{|\mathcal{S}||\mathcal{G}||\mathcal{A}| \cdot \kappa}{(1-\gamma)^3 N}}, \quad \kappa = \sup_{s \in \mathcal{S}, a \in \mathcal{A}, g \in \mathcal{S}} \frac{d^{\pi^\star}(s,a, g)}{d^{\pi^{BC}}(s,a,g)},
    \notag
\end{equation}
with constant probability of $1 - \delta$. Here, the proportionality constant depends on $\delta$. $\kappa$ is the concentrability coefficient that accounts for the distribution shift in data collected by the offline behaviour cloning policy $\pi^{\text{BC}}$, and the data that would have been collected under the optimal policy $\pi^\star$. $d^\pi(s,a,g)$ is the discounted occupancy measure (stationary distribution) of the policy $\pi$. Intuitively, if the dataset does not include the state-action pairs required to learn the optimal policy, $\kappa$ (and hence the error $\epsilon$) will be large. 
\newline
\newline
We build on the work of \citet{robert_sample_nodate} and \citet{li_towards_nodate} to show that, by using a hierarchical policy with absolute abstraction, the maximum error is bounded by: 
\begin{equation}
    \epsilon^{\text{hierarchy, rep}}  \propto \sqrt{\frac{|\mathcal{C}_h| |\Omega| \cdot \kappa^{\text{rep}}_h}{(1-\gamma^n)^3 N}} +  \sqrt{\frac{|\mathcal{C}_l|\mathcal{A}|n^3 \cdot \kappa^{\text{rep}}_l}{N}}. 
    \notag
\end{equation}
This reparameterisation reduces the error bound in four ways:
\begin{enumerate}
    \item \textbf{Horizon reduction}: the high-level policy faces an effective discount factor of $\gamma^n$ (with $n \ge 1$) rather than $\gamma$, reducing the denominator's sensitivity: $\frac{1}{(1 - \gamma^n)^3} \le \frac{1}{(1 - \gamma)^3}$\footnote{Note that while hierarchy introduces an additive error term for the low-level policy, this is typically dominated by the exponential reduction in the high-level error's horizon-dependent constant, especially in tasks where $\gamma \to 1$ \citep{park_horizon_2025}.}.
    \item \textbf{Cardinality Reduction}: by mapping $(s,g)$ and $(s, \omega)$ pairs into equivalence classes $c_h \in \mathcal{C}_h$ and $c_l \in \mathcal{C}_l$, the effective state-space is reduced: $|\mathcal{C}_h| \le |\mathcal{S}||\mathcal{G}| $ and $|\mathcal{C}_l| \le |\mathcal{S}||\Omega|$. 
    \item \textbf{Option Efficiency}: relativised options ensure that the option space is small and invariant to absolute position: $|\Omega^{\text{rel}}| \le |\Omega^{\text{abs}}|$, where $\Omega^{\text{abs}}$ represents an option space anchored in an absolute frame of reference, and $\Omega^{\text{rel}}$ represents one in a relative frame of reference.  
    \item \textbf{Concentrability Improvement}: because the concentrability coefficients are now defined using reparameterised policies over the reparameterised latent spaces, the probability mass of the dataset is aggregated across similar contexts:
    \begin{equation}
        \kappa^{\text{rep}}_h = \sup_{c_h \in \mathcal{C}_h, \omega \in \Omega} \frac{d^{\pi_h^\star}(c_h, \omega)}{d^{\pi_h^{BC}}(c_h,\omega)} \quad \text{and} \quad \kappa^{\text{rep}}_l = \sup_{c_l \in \mathcal{C}_l, a \in \mathcal{A}} \frac{d^{\pi_l^\star}(c_l, a)}{d^{\pi_l^{BC}}(c_l,a)}, 
        \notag
    \end{equation}
    where
    \begin{align}
    d^{\pi_h}(c_h, \omega) = \sum_{(s, g) \in \phi_h^{-1}(c_h)} d^\pi_h(s,g,\omega) \quad \text{and} \quad d^{\pi_l}(c_l, a) = \sum_{(s, \omega) \in 
    \phi_l^{-1}{(c_l)}} d^\pi_l(s,\omega, a). 
    \notag
    \end{align}
    Even if the dataset has zero mass on a specific state-goal-option $(s, g, \omega)$ or state-option-action $(s, \omega, a)$, it likely has mass on the abstract-context-option $(c_h, \omega)$ or abstract-context-action $(c_l, a)$. This reduces the likelihood of a support mismatch, where the optimal policy requires a state transition on which the dataset places zero mass. By aggregating similar contexts we now perform a ratio over sums such that
    \begin{equation}
        \kappa^{\text{rep}}_h \le \kappa_h \quad \text{and} \quad \kappa^{\text{rep}}_l \le \kappa_l, \notag
    \end{equation}
    where $\kappa_h$ and $\kappa_l$ are concentrability coefficients without using aggregating embeddings.
\end{enumerate}
Consequently, for a fixed $N$ and unlike a flat policy, such a reparameterisation could enable learning more optimal behaviour in regions that suffer from low-quality data.
\end{tcolorbox}

\subsection{Abstractive RL Implicitly Learning Relativised Options}
\label{sec: implementation_1}
Based on this objective, we propose a minimal amendment to HIQL \citep{park_hiql_2024} to encourage learning relativised options, which directly address the third point in the Motivation Box (Section \ref{sec: objective}). Rather than learning option representations with the value function, we propose learning them via the low-level policy. Following HIQL, we also bound the option space to a hypersphere to introduce geometric regularisation. However, unlike HIQL, we learn representations via the low-level policy to push together state-waypoint pairs with similar low-level actions rather than state-waypoint pairs with similar values. Also unlike HIQL (and, as motivated in Section \ref{sec: objective}) we use two value functions rather than one. 

\textbf{Low-Level Value.} Although ARL is agnostic to the choice of loss, we train the low-level value $V_l$ and critic $Q_l$ using Implicit Q Learning \citep{kostrikov_offline_2021} to match our benchmark algorithms:
\begin{align}
    \mathcal{L}_{V_l} &= \mathbb{E}_{(s,a) \sim \mathcal{D}, g_s \sim p^\mathcal{D}_{\gamma_l}(\cdot \mid s, a))} \left[\ell^2_\tau\left(V_l(s, g_s) - \tilde{Q}_l(s, g_s,a) \right)\right] \label{eq: low_v} \\
    \mathcal{L}_{Q_l} &= \mathbb{E}_{(s,a,s') \sim \mathcal{D}, g_s \sim p^\mathcal{D}_{\gamma_l}(\cdot \mid s, a))} \left[\left(Q_l(s,g_s,a) - r(s,g_s) - \gamma_l V_l(s', g_s)\right)^2\right] \label{eq: low_q}
\end{align}
where, $\tilde{Q}_l$ denotes the target network, $\gamma_l$ the low-level discount factor, $\ell^2_\tau$ the expectile loss, and $g_s \in \mathcal{S}$ a waypoint to the goal. 

\textbf{Low-Level Policy.} We learn the option embeddings $\phi_\omega$ jointly with the low-level policy. As with the value function, ARL is agnostic to the choice of policy and policy extraction algorithm. In our implementations we use Advantage-Weighted Regression (AWR) \citep{peng_advantage-weighted_2019}:
\begin{equation}
    \begin{aligned}
    \mathcal{L}_{\pi_l, \phi_\omega} &= -\mathbb{E}_{(s,a,s' \cdots g_s) \sim \mathcal{D}} \left[ e^{\alpha_lA_l(s, a, s', g_s)} \log \pi_l\left(a \mid s, \hat\phi_\omega(s, g_s)\right)\right],
    \end{aligned}
    \label{eq: low_awr}
\end{equation}
where $A_l(s, a, s',g_s)=Q_l(s,g_s, a) - V_l(s,g_s)$ represents the advantage associated with action $a$. 

\textbf{High-Level Value.} To stabilise training and mitigate the issue of simultaneously learning the option representation, which can lead to training instability, we learn an action-free high-level value function, which can be learned directly from state trajectories without requiring explicit option labels. Note that, apart from being action-free, ARL is agnostic to the choice of high-level value learning and could be implemented with value horizon reduction such as TD-$n$ or TRL \citep{park_transitive_2026}. In our implementations, we use one-step IVL \citep{ghosh_reinforcement_2023, xu_policyguidedimitationapproachoffline_2023}:
\begin{equation}
    \mathcal{L}_{V_h} = \mathbb{E}_{(s,a) \sim \mathcal{D}, g \sim p^\mathcal{D}(\cdot \mid s, a))} \left[\ell^2_\tau\left(V_h(s,g) - r(s,g) - \gamma \tilde{V}_h(s',g)) \right)\right],
    \label{eq: high_v}
\end{equation}
where $\tilde{V}_h$ represents the high-level target network. Although this biases the high-level value function towards being optimistic in stochastic environments, future work could incorporate a notion of reachability from the low-level value function. 

\textbf{High-Level Policy.} Again, ARL is agnostic to the choice of high-level policy extraction. We hypothesise the high-level policy to be multi-modal, corresponding to distinct and equally optimal options, but note that choice of high-level policy is orthogonal to this work. A high-level Q-function could also be fitted to the high-level value function, which would enable use of Behaviour-Cloned Deep Deterministic Policy Gradient (DDPGBC) for policy extraction \citep{fujimoto_minimalist_2021}, for example; we include details in Appendix \ref{app: algos}. In our implementations, we use a Gaussian high-level policy, which we generally (see Appendix \ref{app: details}) train using AWR:
\begin{equation}
    \begin{aligned}
    \mathcal{L}_{\pi_h} &= -\mathbb{E}_{(s \cdots g_s) \sim \mathcal{D}, g \sim p^\mathcal{D}_\gamma(\cdot \mid s)} \left[ e^{\alpha_hA_h(s, \hat\phi_\omega(s,g_s), g_s, g)} \log \pi_h\left(\hat\phi_\omega(s, g_s) \mid s, g \right)\right],
    \label{eq: high_awr}
    \end{aligned}
\end{equation}
where $A_h$ represents the advantage associated with option $\hat\phi_\omega(s, g_s)$.
We provide pseudocode in Algorithm \ref{alg: arli} in Appendix \ref{app: algos}. Full experiment details, hyperparameters, sampling methods and seeds are found in Appendices \ref{app: algos} and \ref{app: details}, and in our codebase.

\subsection{Abstractive RL Explicitly Enforcing Translation Invariance}
\label{sec: implementation_2}
We now introduce a second algorithm, which explicitly imposes translation invariance on state-waypoint representations in the low-level decision process in order to learn from similar contexts across the state-space. We hypothesise that such an inductive bias could be useful in manipulation tasks, for example. We propose this algorithm as a proof of concept that using different representations at different levels of the hierarchy can improve generalisation in Offline RL. 

We define \emph{relativised states} as an unnormalised displacement vector: 
\begin{equation}
    v = g_s - s, \notag
\end{equation}
resulting in a single vector that simultaneously encodes both the state and waypoint.

Since hard-coding representations can impose representational constraints, our approach exploits the two-step procedure of Offline RL as a compromise. To enforce experience reuse, we define the low-level value function in terms of relativised states: $V_l(g_s - s)$. Although this introduces a representational constraint (where state-waypoint pairs with distinct values may map to the same relative vector\footnote{For instance, in a maze, a state $s$ and waypoint $g_s$ separated by a wall may map to the same relative vector as a pair in open space, yet induce vastly different value estimates.}), it explicitly collapses the state-waypoint space into a manifold of relative displacements, addressing the second and fourth points in the Motivation Box (Section \ref{sec: objective}). We remark that the issue of representational constraints could also be mitigated by computing the difference of encoded representations such that  $v = \phi_{l_{g_s}}(g_s)-\phi_{l_s}(s)$, although we did not find this to be necessary to achieve superior performance in our experiments. 

As in Section \ref{sec: implementation_1}, to relativise options and address the third point in the Motivation Box, option-embeddings are learned with the low-level policy. However, now options are also explicitly relativised by defining them in terms of relativised states. We use soft-normalisation rather than length-normalisation to avoid numerical instability while introduce geometric regularisation and allow re-normalising of samples from the high-level policy \citep{park_hiql_2024} upon deployment: 
\begin{equation}
    \begin{aligned}
        o := \hat\phi_\omega(s, g_s) = \frac{\phi_\omega(g_s - s) \cdot \tanh{||\phi_\omega(g_s - s)||}}{||\phi_\omega(g_s - s)||} \cdot \sqrt{d},
    \notag
    \end{aligned}
\end{equation}
where $d$ is the dimension of the embedding $\phi_\omega$, and $||\cdot||$ denotes the standard Euclidean norm. 
Soft normalisation means that the option space includes the space within the hypersphere and allows options to incorporate implicit temporal awareness as their magnitude scales linearly when displacement is small.  

To satisfy local constraints, we still condition the low-level policy on the absolute state: $\pi_l(\cdot \mid s, \hat{\phi}_\omega(s, g_s))$. We provide pseudocode in Algorithm \ref{alg: arle} in Appendix \ref{app: algos}. Full experiment details, hyperparameters, sampling methods and seeds are found in Appendices \ref{app: algos} and \ref{app: details} and in our codebase.

\section{Experiments}
\label{sec: experiments}
The goal of our experiments is simple: to test whether relativised options and distinct representations at different levels of the hierarchy can lead to better policy generalisation in offline GCRL. 

\paragraph{Benchmark and Ablations.} We perform all experiments on the standard OGBench datasets, focusing on the more challenging locomotion and manipulation environments (i.e. selecting \textit{giant} over \textit{medium} or \textit{large}). For completeness, we also evaluate on a stochastic setting (\textit{teleport}), despite its mismatch with our deterministic assumption (Section \ref{sec: prelims}). We exclude visual environments as they introduce additional challenges related to high-dimensional perception that are orthogonal to this work. Unlike prior work \citep{park_horizon_2025, park_transitive_2026}, we do not use oracle representations, which simplify option learning in locomotion, and discard proprioceptive information that might be useful in manipulation. 

\begin{figure}
    \centering
    \includegraphics[width=1.0\linewidth]{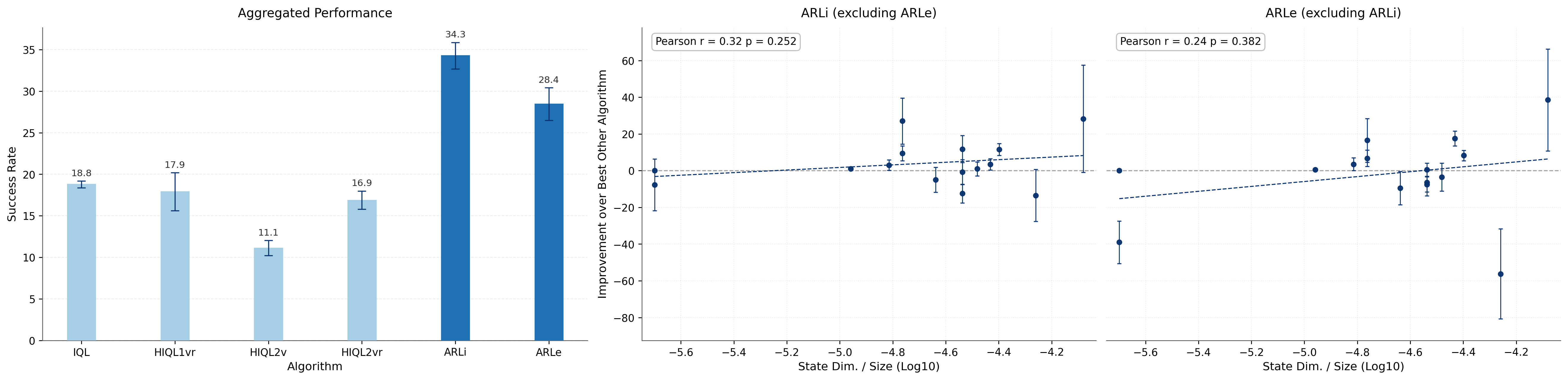}
    \caption{\textbf{Analysis.} Aggregate Performance across all tasks (\textbf{left}) and ARLi's (\textbf{middle}) and ARLe's (\textbf{right}) performance improvements over next-best performing algorithm against number of state dimensions per dataset sample. Bootstrapped 95\% CI over 4 seeds and 20 evaluation runs.}
    \label{fig: analysis}
\end{figure}

We benchmark ARL with implicitly learned relativised options (Section \ref{sec: implementation_1}, \textbf{ARLi}), and explicitly enforced translation invariance (Section \ref{sec: implementation_2}, \textbf{ARLe}), against the original version of HIQL \citep{park_hiql_2024} (\textbf{HIQL1vr}), which uses a single value function and learns option representations via this value function. To isolate the effect of the relativised representation rather than any differences arising due to structure of the value function (ARL employs two value functions), we compare against variants of HIQL that also use two value functions. We include a variant with two value functions that does not include option representation learning (\textbf{HIQL2v}), and a variant with two value functions that learns option representations via the low-level value function (\textbf{HIQL2vr}), with the intention of mirroring HIQL1vr. Finally, we also compare against goal-conditioned IQL \citep{kostrikov_offline_2021}, the best-performing flat policy from \citet{park_ogbench_2025} (\textbf{IQL}). We refer to Appendix \ref{app: algos} and our codebase for full implementation details.

Since hyperparameter tuning is expensive and ARL, which is based on inductive biases, introduces no additional hyperparameters over HIQL, we simply adopt those tuned for HIQL \citep{park_ogbench_2025, park_horizon_2025} (see Appendix \ref{app: details}). The fact that ARL achieves strong performance under these hyperparameters attests to its efficacy. ARL is agnostic to the choice of policy class and value learning objective, so, to avoid related confounding factors, we use a Gaussian and one-step TD for all experiments. 

\textbf{Results.} Our results (Table \ref{tab: results} and left of Figure \ref{fig: analysis}) show that both variants of ARL outperform both the flat and absolute hierarchical policies, achieving a mean success rate that is at least 10 percentage points higher than the benchmarks when aggregating over all tasks. We highlight that this is without necessitating hyperparameter tuning.

Simply by learning relativised options based on action similarity rather than value similarity, ARLi consistently improves over the hierarchical benchmarks (excluding ARLe). In particular, it more than doubles the success rate in both the \textit{humanoidmaze} environments, and almost doubles it for \textit{scene-play-v0}.

ARLe performs especially well in the manipulation environments. Most notably, in \textit{puzzle-4x4-play-v0}, an 83-dimensional manipulation task, ARLe achieves an 88\% success rate, outperforming the next-best benchmark by 39 percentage points (excluding ARLi). We hypothesise that translational invariance is particularly beneficial in high-dimensional settings with underlying symmetries and sparse state-space coverage. To investigate this, we plot improvement over the next-best benchmark\footnote{We try to mitigate confounding factors such as horizon length, absolute dataset size and policy expressivity (e.g. whether unimodal or multimodal) by plotting performance gains over the next-best algorithm rather than absolute success rate.} against state dimensionality normalised by dataset size (right of Figure \ref{fig: analysis}). Although correlation does not imply causation, and the observed correlations are weak and not statistically significant (amplified by a small number of environments\footnote{Note that this is a common issue in Deep RL, and, like prior work \citep{agarwal2021deep}, we emphasise that lack of statistically significant results does not demonstrate the absence of effect.}), both ARLi and ARLe exhibit positive improvement trends with increasing dimensional sparsity. In comparison, HIQL1vr and HIQL2vr, for example, show stronger negative trends with increasing sparsity (Figure \ref{fig: full_correlation} in Appendix \ref{app: results}) under the same methodology. Intuitively, it makes sense that relativised options and experience reuse become increasingly important as sparsity increases.  
\begin{table}[t]
\centering
\setlength{\tabcolsep}{5.5pt}
\caption{\textbf{Results.} We report each method's average (binary) success rate (\%) across the five test-time goals. Bootstrapped 95\% CI over 4 seeds and 20 evaluation runs. Blue bold indicates the highest mean; black bold overlapping confidence intervals.}
\footnotesize
\begin{tabular}{lcccccccc}
\toprule
\textbf{Task} & \textbf{Size} & \textbf{Dim.} & \textbf{IQL} & \textbf{HIQL1vr} & \textbf{HIQL2v} & \textbf{HIQL2vr} & \textbf{ARLi} & \textbf{ARLe} \\
\midrule
pointmaze-giant-navigate-v0 & 1M & 2 & 0$\pm$0 & \textcolor{clearsky!95!black}{\textbf{55$\pm$11}} & 21$\pm$7 & \textbf{46$\pm$6} & \textbf{48$\pm$8} & 16$\pm$2\\
pointmaze-giant-stitch-v0 & 1M & 2 & 0$\pm$0 & \textbf{0$\pm$0} & 0$\pm$0 & 0$\pm$0 & 0$\pm$0 & 0$\pm$0\\
\midrule
antmaze-giant-navigate-v0 & 1M & 29 & 0$\pm$0 & 38$\pm$3 & \textbf{40$\pm$3} & \textcolor{clearsky!95!black}{\textbf{48$\pm$6}} & \textcolor{clearsky!95!black}{\textbf{48$\pm$3}} & \textbf{42$\pm$4}\\
antmaze-giant-stitch-v0 & 1M & 29 & 0$\pm$0 & 7$\pm$7 & 15$\pm$3 & 20$\pm$3 & \textcolor{clearsky!95!black}{\textbf{32$\pm$7}} & 21$\pm$1\\
antmaze-teleport-stitch-v0 & 1M & 29 & \textcolor{clearsky!95!black}{\textbf{49$\pm$2}} & 29$\pm$4 & 43$\pm$3 & 40$\pm$6 & 36$\pm$5 & 41$\pm$4\\
\midrule
humanoidmaze-giant-navigate-v0 & 4M & 69 & 1$\pm$1 & 22$\pm$11 & 12$\pm$5 & 11$\pm$10 & \textcolor{clearsky!95!black}{\textbf{49$\pm$6}} & 38$\pm$4\\
humanoidmaze-giant-stitch-v0 & 4M & 69 & 0$\pm$0 & 4$\pm$3 & 0$\pm$0 & 0$\pm$0 & \textcolor{clearsky!95!black}{\textbf{13$\pm$2}} & \textbf{10$\pm$3}\\
\midrule
cube-double-play-v0 & 1M & 37 & 50$\pm$3 & 2$\pm$0 & 0$\pm$0 & 3$\pm$0 & 53$\pm$2 & \textcolor{clearsky!95!black}{\textbf{67$\pm$3}}\\
cube-triple-play-v0 & 3M & 46 & \textbf{11$\pm$2} & 7$\pm$3 & 0$\pm$0 & 1$\pm$1 & \textbf{14$\pm$2} & \textcolor{clearsky!95!black}{\textbf{15$\pm$3}}\\
cube-quadruple-play-v0 & 5M & 55 & 0$\pm$0 & 0$\pm$0 & 0$\pm$0 & 0$\pm$0 & \textcolor{clearsky!95!black}{\textbf{1$\pm$0}} & 0$\pm$0\\
\midrule
puzzle-3x3-play-v0 & 1M & 55 & \textcolor{clearsky!95!black}{\textbf{100$\pm$0}} & 27$\pm$8 & 0$\pm$0 & 24$\pm$3 & \textbf{86$\pm$14} & 44$\pm$25\\
puzzle-4x4-play-v0 & 1M & 83 & 30$\pm$3 & 49$\pm$27 & 0$\pm$0 & 17$\pm$6 & \textbf{78$\pm$11} & \textcolor{clearsky!95!black}{\textbf{88$\pm$6}}\\
puzzle-4x5-play-v0 & 3M & 99 & \textbf{15$\pm$3} & \textbf{17$\pm$3} & 0$\pm$0 & \textbf{12$\pm$3} & \textcolor{clearsky!95!black}{\textbf{18$\pm$3}} & \textbf{14$\pm$7}\\
puzzle-4x6-play-v0 & 5M & 115 & 13$\pm$1 & 0$\pm$0 & 0$\pm$0 & \textcolor{clearsky!95!black}{\textbf{18$\pm$2}} & \textbf{14$\pm$7} & \textbf{9$\pm$9}\\
\midrule
scene-play-v0 & 1M & 40 & 13$\pm$2 & 12$\pm$3 & 12$\pm$7 & 13$\pm$1 & \textcolor{clearsky!95!black}{\textbf{24$\pm$3}} & \textbf{21$\pm$3}\\
\bottomrule
\end{tabular}
\label{tab: results}
\end{table}
\begin{wrapfigure}[13]{r}{0.6\linewidth}
    \centering
    \includegraphics[width=1.0\linewidth]{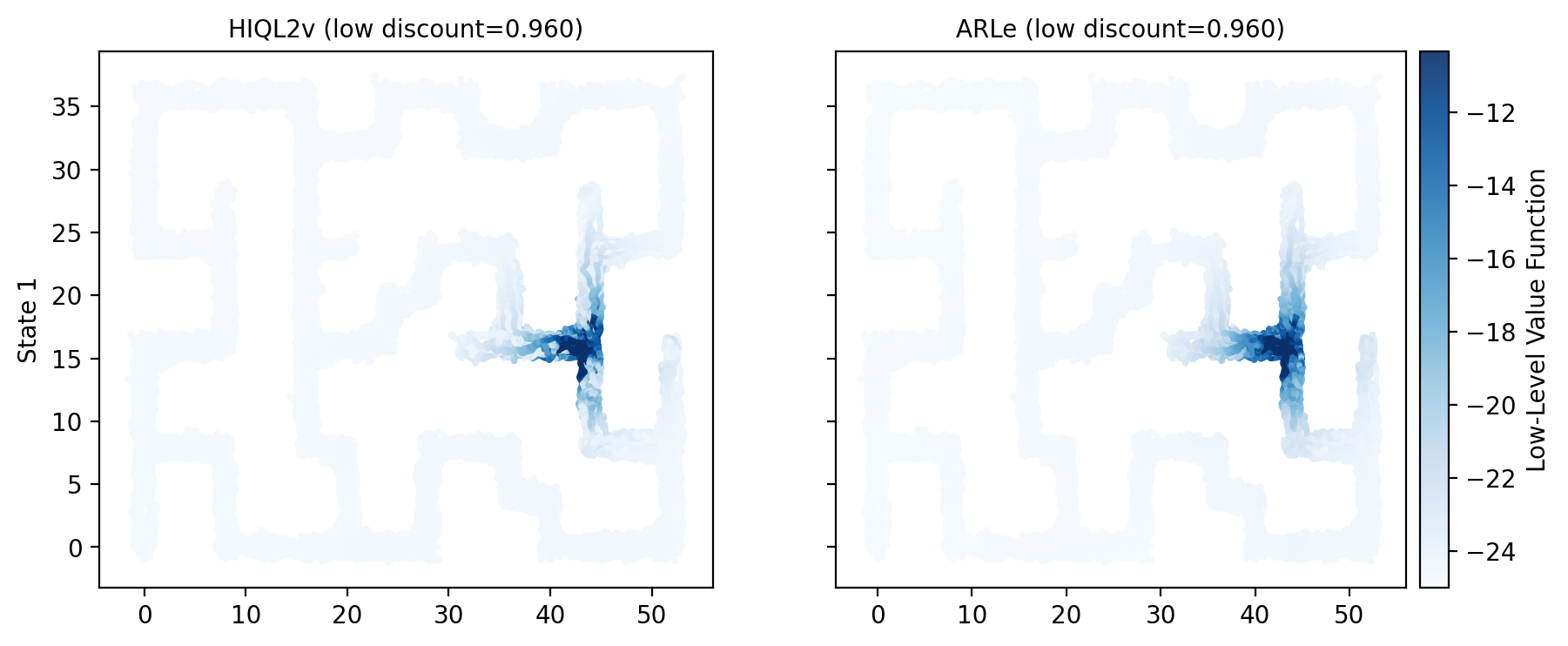}
    \caption{\textbf{Low-level} value function for \textbf{HIQL2v} (\textbf{left}) and \textbf{ARLe} (\textbf{right}) (task 4, \textit{antmaze-giant-stitch-v0}).}
    \label{fig: low_value}
\end{wrapfigure}
To better understand the effect of imposing translational invariance, we visualise the low-level value functions for ARLe and HIQL2v in the \textit{antmaze} locomotion environment (Figure \ref{fig: low_value}). Due to explicitly collapsing equivalent relative states, ARLe learns a substantially smoother low-level value function. We now address potential questions.

\textbf{Why not tune hyperparameters for ARL?} Offline RL typically requires significant online tuning \citep{jackson_clean_2025}, which is expensive and would be unscalable for training a billion-parameter general-purpose agent \citep{kaplan_scaling_2020, wang_1000_2026}. By using inductive biases rather than representational losses, we avoid introducing additional hyperparameters: ARL achieving significant performance gains under hyperparameters tuned for HIQL indicates robustness. 

\textbf{Why introduce ARLe if ARLi is so consistent?} 
ARLe demonstrates how hierarchical structures and relativised options enable level-specific representations, leveraging this inductive bias to improve performance by 10 percentage points over ARLi in two out of the four sparsest datasets (\textit{puzzle-4x4-play-v0} and \textit{cube-double-play-v0}).  

\textbf{Why does ARLe perform poorly in certain tasks?} 
ARLe’s performance depends on the alignment between its inductive bias and the environment's structure. Assuming local translational invariance can benefit high-dimensional manipulation through experience reuse but can be detrimental in dense, low-dimensional environments like \textit{pointmaze}, since the representation is inherently lossy. We also hypothesise that mapping 2D relative displacements into a 10D latent hypersphere introduces representational noise. While future work could leverage GDL to learn more flexible symmetries, ARLe demonstrates that decoupling representations across the hierarchy enables a degree of experience reuse fundamentally inaccessible to flat or absolute-frame architectures. When the inductive bias is well-matched to the environment, it significantly enhances policy generalisation.

\textbf{Why do all algorithms perform poorly in certain tasks?}
We hypothesise this to be due to other confounding factors such as: (i) poor high-level value learning and a lack of gradient in long horizon tasks (we include plots of the high-level value function in Figure \ref{fig: high_level_value} in Appendix \ref{app: results}); (ii) the high-level policy being unimodal rather than multimodal; (iii) not training for enough gradient steps for large dataset sizes; (iv) not learning goal representations for the high-level policy. The aforementioned issues could be mitigated by combining ARL with value horizon reduction methods such as TD-$n$ or TRL \cite{park_transitive_2026}, learning a flow-policy \cite{lipman_flow_2024} rather than a Gaussian (especially for the high-level policy), training for more steps (we run all experiments for 1M), and using, for example, Dual-Goal Representations \citep{park_dual_2026} for the high-level decision process. 

\section{Conclusion}
\label{sec: conclusion}
In this work we motivate hierarchy in offline RL through \emph{absolute} abstraction. By learning relativised options and using distinct representations at different levels of the hierarchy, agents can reuse optimal experience across similar contexts of the state-space, enabling better performance in regions of the dataset only supported by low-quality data. Based on our framework, we introduce two simple algorithms for learning relativised options via action similarity and explicitly enforcing translational invariance on the low-level decision process. Our experiments demonstrate that such relativised options and inductive biases improve policy generalisation in high-dimensional offline GCRL. This proof of concept opens many avenues for future research, including imposing more flexible inductive biases, or leveraging action-chunking to learn relativised options over action sequences. We hope that this work motivates progress towards scalable offline RL.
\newpage

\bibliography{arl}
\bibliographystyle{unsrtnat}

\newpage

\appendix
\section{Sample Complexity in Online Goal-Conditioned RL}
We provide some intuition into the choice of policy learning and representation using sample complexity in finite state and action space problems. We build on the works of \citep{kakade2003sample, munos_finite-time_nodate, lattimore_pac_2012, robert_sample_nodate,  li_towards_nodate}.

In online GCRL, the aim is to find the optimal policy with the smallest number of samples or online interactions, $N$, such that the error in optimal return is smaller than a fixed constant $\epsilon$.

\subsection{GCRL Sample Complexity}
Consider the case of a discrete, discounted horizon MDP with a finite state space $\mathcal{S}$, action space $\mathcal{A}$ and discount factor $\gamma \in [0, 1)$. The Probably-Approximately Correct (PAC) Learning \cite{kakade2003sample} upper-bound sample complexity to find an $\epsilon$ optimal policy reaching a unique goal-state optimally (assuming at most two possible next-states for each state/action pair) with constant probability of $1 - \delta$ is given by \cite{lattimore_pac_2012}: 
\begin{equation}
    N^{\text{infinite, single goal}} \propto \frac{|\mathcal{S}||\mathcal{A}|}{(1-\gamma)^3\epsilon^2 }. 
    \notag
\end{equation}
Hence, the minimax sample complexity required to find an $\epsilon$ optimal policy reaching any given state (such that we have $|\mathcal{G}|$ unique goal-states) is given by: 
\begin{equation}
    N^{\text{infinite}} \propto \frac{|\mathcal{S}||\mathcal{G}||\mathcal{A}|}{(1-\gamma)^3\epsilon^2 }. 
    \label{eq: gc_sc_infinite}
\end{equation}

Consider the case of a discrete finite horizon (of length $H$) MDP with a finite state space $\mathcal{S}$, action space $\mathcal{A}$ and discount factor $\gamma \in [0, 1)$. The minimax sample complexity to find an $\epsilon$ optimal policy reaching a unique goal-state optimally is given by: 
\begin{equation}
    N^{\text{finite}} \propto \frac{|\mathcal{S}||\mathcal{G}||\mathcal{A}|H^3}{\epsilon^2 }. 
    \notag
\end{equation}

\subsection{GCRL Hierarchical Sample Complexity}
Using a hierarchical policy, we can break the distant goal $g$ from our current state $s$ into options defined over an option space $\Omega$. Hence, the high-level policy becomes
\begin{equation}
    N^{\text{high}} \propto \frac{|\mathcal{S}||\mathcal{G}||\Omega|}{(1-\gamma^n)^3\epsilon^2 },
    \notag
\end{equation}
where we have substituted the option-space to Equation \ref{eq: gc_sc_infinite}, and use the environment's discount factor raised to a factor of $n$, assuming that the high-level policy acts, on average, every $n$ steps.

The low-level policy has a sample complexity of 
\begin{equation}
    N^{\text{low}} \propto \frac{|\mathcal{S}||\Omega||\mathcal{A}|n^3}{\epsilon^2 }, 
    \notag
\end{equation}
since $|\Omega|$ options can be executed. 
Hence, the overall sample complexity of the hierarchical policy $\pi(a \mid s,g) := \pi_l(a \mid s, \omega), \quad \omega \sim \pi_h(\cdot \mid s,g)$ is \citep{robert_sample_nodate}: 
\begin{equation}
    N^{\text{hierarchy}} \propto \frac{|\mathcal{S}||\mathcal{G}||\Omega|}{(1-\gamma^n)^3\epsilon^2} + \frac{|\mathcal{S}||\Omega||\mathcal{A}|n^3}{\epsilon^2 }. 
    \label{eq: sc_hrl}
\end{equation}

Note, that, in the case of no temporal abstraction, when $n = 1$, we approximately recover the original sample complexity of a flat policy. Since then the option just becomes a single primitive action, Equation \ref{eq: sc_hrl} becomes:
\begin{equation}
    N^{\text{hierarchy}} \propto \frac{|\mathcal{S}||\mathcal{G}||\mathcal{A}|}{(1-\gamma)^3\epsilon^2} + \frac{|\mathcal{S}||\mathcal{A}|^2}{\epsilon^2 }
    \approx \frac{|\mathcal{S}||\mathcal{G}||\mathcal{A}|}{(1-\gamma)^3\epsilon^2},
    \notag
\end{equation}
where the approximation follows due to the dominating $\frac{1}{(1 - \gamma)^3}$ factor in the first term, assuming long-horizon problems such that $\gamma \to 1$ and that the goal-space is larger or equal to the size of the action space $|\mathcal{G}| \ge |\mathcal{A}|$ \footnote{Even though this might not be the case, usually the goal-space is unknown, so we train the policy to reach any state within the state-space i.e. such that $\mathcal{G} = \mathcal{S}$. Assuming that $|\mathcal{S}| \gg |\mathcal{A}|$ is a standard assumption.}. 

Issues arise under a misspecified option horizon $n$. In this case, the sample complexity of the hierarchical policy becomes worse than that of a flat policy, as the skill space must then account for every sequence of primitive actions over $n$ steps, such that $|\Omega| = |\mathcal{A}|$. The sample complexity of the hierarchical policy becomes
\begin{equation}
    N^{\text{hierarchy}} \propto \frac{|\mathcal{S}||\mathcal{G}||\mathcal{A}|^n}{(1-\gamma^n)^3\epsilon^2} + \frac{|\mathcal{S}||\Omega||\mathcal{A}|^nn^3}{\epsilon^2 },
    \notag
\end{equation}
which blows up due to the exponential factor multiplying the action space. 

\subsection{GCRL State Representation Sample Complexity}
State representation sample complexity exploits symmetry in the state-goal space to a mapping $\phi(s,g)\to c$ such that $\phi(s,g) = \phi(s',g')$ if $\pi^\star(a \mid s,g) = \pi^\star(a \mid s',g')$. The sample complexity of learning an optimal policy (Equation \ref{eq: gc_sc_infinite}) becomes: 
\begin{equation}
    N^{\text{rep}} \propto \frac{|\mathcal{C}||\mathcal{A}|}{(1-\gamma)^3\epsilon^2}. 
    \label{eq: gc_sc_sr}
\end{equation}
Note that $|\mathcal{C}| \le |\mathcal{S}||\mathcal{G}|$, with $|\mathcal{C}| = |\mathcal{S}||\mathcal{G}|$ if each $(s,g)$ has a distinct optimal action. 

\section{Error in Offline Goal-Conditioned RL}
\label{app: error}
In offline GCRL, the aim is to bound the error $\epsilon$ given an offline dataset $\mathcal{D}$ of fixed size $N$.  

Unlike in the online setting, where it is assumed that the agent can sample any state-action pair to learn the environment's dynamics, in offline RL, a concentrability coefficient $\kappa$ is incorporated to account for the distribution shift in data collected by the policy $\pi^{\text{BC}}$, and the data that would have been collected induced under the optimal policy $\pi^\star$. Intuitively, if the dataset does not include the states required to learn the optimal policy, the algorithm may never learn that optimal policy. 
The concentrability coefficient $\kappa$ is defined as: 
\begin{equation}
    \kappa = \sup_{s \in \mathcal{S}, a \in \mathcal{A}, g \in \mathcal{G}} \frac{d^{\pi^\star}(s,a, g)}{d^{\pi^{\text{BC}}}(s,a,g)},
    \notag
\end{equation}
where $d^\pi(s,a,g)$ is the discounted occupancy measure (i.e. stationary distribution) of the policy $\pi$:
\begin{equation}
    d^\pi(s,a,g) = (1-\gamma) \mathbb{E}_{\tau \sim\pi, s_0, g_0 \sim \text{Unif}(\mathcal{S})} \left[ \sum_{t=0}^{\infty} \gamma^t \mathbbm{1}(s_t = s, a_t = a, g_0 = g) \right],
    \notag
\end{equation}
and where the trajectory is induced by the MDP and following the policy $\pi( a \mid s, g)$. If the dataset is highly exploratory and covers the optimal paths well, $\kappa$ will be small. If the dataset is narrow, or misses critical regions of the state-action space, $\kappa$ will be large. 

Rearranging Equation \ref{eq: gc_sc_infinite} and incorporating the concentrability coefficient, the offline bound for a goal-conditioned flat policy is given by: 
\begin{equation}
    \epsilon \propto \sqrt{\frac{|\mathcal{S}||\mathcal{G}||\mathcal{A}| \cdot \kappa}{(1-\gamma)^3 N}}.
    \notag
\end{equation}

\subsection{GCRL Hierarchical Error}
Since the size of the offline dataset is fixed, the only way to reduce the error $\epsilon$ is to use a hierarchical policy with distinct state-goal representations. 

Using a hierarchical policy, the error term becomes: 
\begin{equation}
    \epsilon^{\text{hierarchy}}  \propto \sqrt{\frac{|\mathcal{S}||\mathcal{G}||\Omega| \cdot \kappa_h}{(1-\gamma^n)^3 N}} +  \sqrt{\frac{|\mathcal{S}||\Omega||\mathcal{A}|n^3 \cdot \kappa_l}{N}},
    \notag
\end{equation}
where the concentrability coefficients are defined as: 
\begin{equation}
    \kappa_h = \sup_{s \in \mathcal{S}, \omega \in \Omega, g \in \mathcal{S}} \frac{d^{\pi_h^\star}(s,\omega, g)}{d^{\pi_h^{\text{BC}}}(s,\omega,g)} \quad \text{and} \quad \kappa_l = \sup_{s \in \mathcal{S}, a \in \mathcal{A}, \omega \in \Omega} \frac{d^{\pi_l^\star}(s,a, \omega)}{d^{\pi_l^{\text{BC}}}(s,a,\omega)}.
    \notag
\end{equation}

Then, introducing two embeddings that group together state-goal pairs $(s,g)$ requiring similar options $\omega$ such that $\phi_h(s,g) = \phi_h(s',g') = c_h$ if $\pi^\star_h(\cdot \mid s,g) = \pi^\star_l(\cdot \mid s',g')$ and $c_h \in \mathcal{C}_h$, and state-option $(s,\omega)$ requiring similar low-level actions $\phi_l(s,\omega) = \phi_l(s',\omega') = c_l$ if $\pi^\star_l(\cdot \mid s,\omega) = \pi^\star_l(\cdot \mid s',\omega')$ \footnote{Note that we should not use these same embeddings for the value or critic functions, since even though $\pi^\star_l(\cdot \mid s,\omega) = \pi^\star_l(\cdot \mid s',\omega')$, this does not instantly imply that $Q_h(s,g,\omega)=Q_h(s',g',\omega')$: even though the ordering over actions might be the same, there might be an offset such that $Q_h(s,g,\omega)=Q_h(s',g',\omega) + c(s,g,s',g') \quad \forall \quad \omega$. We refer the reader to \citet{li_towards_nodate}.} and $c_l \in \mathcal{C}_l$, the error term is bounded by: 
\begin{equation}
    \epsilon^{\text{hierarchy, rep}}  \propto \sqrt{\frac{|\mathcal{C}_h| |\Omega| \cdot \kappa^{\text{rep}}_h}{(1-\gamma^n)^3 N}} +  \sqrt{\frac{|\mathcal{C}_l||\mathcal{A}|n^3 \cdot \kappa^{\text{rep}}_l}{N}}. 
    \notag
\end{equation}
By definition, $|\mathcal{C}_h||\Omega| \le |\mathcal{S}||\mathcal{G}||\Omega|$ and $|\mathcal{C}_l||\mathcal{A}|\le |\mathcal{S}||\Omega||\mathcal{A}|$.

Because the concentrability coefficients are now defined using the reparameterised policies over the reparameterised latent spaces $\mathcal{C}_h$ and $\mathcal{C}_l$, i.e.
\begin{equation}
    \kappa^{\text{rep}}_h = \sup_{c_h \in \mathcal{C}_h, \omega \in \Omega} \frac{d^{\pi_h^\star}(c_h, \omega)}{d^{\pi_h^{\text{BC}}}(c_h,\omega)} \quad \text{and} \quad \kappa^{\text{rep}}_l = \sup_{c_l \in \mathcal{C}_l, a \in \mathcal{A}} \frac{d^{\pi_l^\star}(c_l, a)}{d^{\pi_l^{\text{BC}}}(c_l,a)}, 
    \notag
\end{equation}
the probability mass of the offline dataset is aggregated across state-goal or state-option pairs that are equivalent under these embeddings:
\begin{align}
    d^{\pi_h}(c_h, \omega) = \sum_{(s, g) \in \phi_h^{-1}(c_h)} d^\pi_h(s,g,\omega) \quad \text{and} \quad d^{\pi_l}(c_l, a) = \sum_{(s, \omega) \in 
    \phi_l^{-1}{(c_l)}} d^\pi_l(s,\omega, a). 
    \notag
\end{align}
This reduces the likelihood of a support mismatch, where the optimal policy requires a state transition on which the dataset places zero mass. Hence, 
\begin{equation}
    \kappa^{\text{rep}}_h \le \kappa_h \quad \text{and} \quad \kappa^{\text{rep}}_l \le \kappa_l. \notag
\end{equation}

\newpage
\section{Offline RL Algorithms}
\label{app: algos}

In the following section $g_s$ is a waypoint to the goal, such that $g_s \in \mathcal{S}$. When sampled from the dataset $\mathcal{D}$, $g_s$ is $n$ steps ahead of the current state $s$. When sampling the waypoint $g_s$ or goal $g$ from $p^\mathcal{D}(\cdot \mid s,a)$, we sample either from a geometric distribution according to the specified discount factor, or a uniform distribution. Details are given in Appendix \ref{app: details}. 

\subsection{Implicit Q-Learning (IQL)} 
\label{app: iql}
The flat policy we benchmark is IQL \citep{kostrikov_offline_2021}, which trains a state-goal value function $V(s,g)$ and state-goal-action value function $Q(s,g,a)$ using the following losses: 
\begin{equation}
    \mathcal{L}_V = \mathbb{E}_{(s,a) \sim \mathcal{D}, g \sim p^\mathcal{D}(\cdot \mid s, a))} \left[\ell^2_\tau\left(V(s,g) - \tilde{Q}(s,g,a) \right)\right],
    \notag
\end{equation}
where $\tilde{Q}$ denotes the target network, and $\ell^2_\tau$ denotes the expectile loss $\ell^2_\tau(x) = |\tau - (x < 0)|x^2$, and 
\begin{equation}
    \mathcal{L}_Q = \mathbb{E}_{(s,a,s') \sim \mathcal{D}, g \sim p^\mathcal{D}(\cdot \mid s, a))} \left[\left(Q(s,a,g) - r(s,g) - \gamma V(s',g)\right)^2. \right]
    \notag
\end{equation}
A Gaussian policy is then extracted using the following DDPGBC \citep{fujimoto_minimalist_2021} loss: 
\begin{equation}
    \mathcal{L}^{\text{DDPGBC}}_\pi = -\mathbb{E}_{(s,a, s') \sim \mathcal{D}, g \sim p^\mathcal{D}_\gamma(\cdot \mid s, a)}\left[Q(s, g, \mu^{\pi}(s,g)) + \alpha \log \pi(a \mid s,g) \right],
\end{equation}
which has been found to outperform AWR \citep{park_is_2024}. 

\subsection{Hierarchical Implicit Q-Learning 1 Value Function with Representation Learning (HIQL1vr)}
\label{app: hiql1vr}
We benchmark against HIQL \citep{park_hiql_2024}, which trains a single, action-free state-goal value function $V(s,g)$ using Implicit V-Learning (IVL) \citep{park_ogbench_2025} and extracts hierarchical policies using AWR-like objectives. The parameterisation $\phi_\omega$ for the low-level policy is learned with the value function $V(s, \hat\phi_\omega(s,g))$, where $\phi_\omega(s,g) \in \mathbb{R}^d$, and normalised such that $|| \hat\phi_\omega (s,g)||^2_2 = d$, where $d$ is the dimension of the embedding. The IVL loss is given by: 
\begin{equation}
    \mathcal{L}_{V, \phi_\omega} = \mathbb{E}_{(s,a, s') \sim \mathcal{D}, g \sim p^\mathcal{D}(\cdot \mid s, a))} \left[\ell^2_\tau\left(V(s,\hat\phi_\omega(s,g)) - r(s,g) - \gamma \tilde{V}(s',\tilde{\hat\phi}_\omega(s',g)) \right)\right],
    \notag
\end{equation}
where $\tilde{V}$ and $\tilde{\hat{\phi}}_o$ denote the target network and representations and $\ell^2_\tau$ the expectile loss, as before. 
The low-level and high-level policies are extracted as follows: 
\begin{equation}
    \mathcal{L}_{\pi_h} = -\mathbb{E}_{(s \cdots g_s) \sim \mathcal{D}, g \sim p^\mathcal{D}_\gamma(\cdot \mid s, a))} \left[ e^{\alpha_h(V(g_s, \hat\phi_\omega(g_s, g)) - V(s, \hat\phi_\omega(s,g))} \log \pi_h\left(\hat\phi_\omega(s,g_s) \mid s,g \right)\right],
    \notag
\end{equation}
\begin{equation}
    \mathcal{L}_{\pi_l} = -\mathbb{E}_{(s,a,s' \cdots g_s) \sim \mathcal{D}} \left[ e^{\alpha_l(V(s', \hat\phi_\omega(s', g_s)) - V(s, \hat\phi_\omega(s,g_s)))} \log \pi_l\left(a \mid s, \hat\phi_\omega(s,g_s) \right)\right].
    \notag
\end{equation}
Since \citet{park_is_2024} found that DDPGBC is better at extracting a policy than AWR, and similarly to \citet{park_hiql_2024}'s action-free value function, we then fit a sort of high-level, action-free Q function to the value function:
\begin{equation}
    \mathcal{L}_{Q_h} = \mathbb{E}_{(s,\cdots, g_s) \sim \mathcal{D}, g \sim p^\mathcal{D}(\cdot \mid s, a))} \left[\left(Q_h(s,g, g_s) - V_h(g_s,g)\right)^2 \right].
    \notag
\end{equation}
This is simply to allow some extrapolation during extraction of the high-level policy.

\subsection{Hierarchical Implicit Q-Learning 2 Value Functions (HIQL2v)}
\label{app: hiql2v}
Since ARL uses two value functions, we also train a second style of HIQL, where we use a low-level and high-level value function. The high-level value function is trained using IVL, while the low-level value function is trained using IQL: 
\begin{equation}
    \mathcal{L}_{V_h} = \mathbb{E}_{(s,a) \sim \mathcal{D}, g \sim p^\mathcal{D}(\cdot \mid s, a))} \left[\ell^2_\tau\left(V_h(s,g) - r(s,g) - \gamma \tilde{V}_h(s',g)) \right)\right],
    \notag
\end{equation}
and 
\begin{equation}
    \begin{aligned}
    \mathcal{L}_{V_l} &= \mathbb{E}_{(s,a) \sim \mathcal{D}, g_s \sim p^\mathcal{D}_{\gamma_l}(\cdot \mid s, a))} \left[\ell^2_\tau\left(V_l(s,g_s) - \tilde{Q}_l(s,g_s,a) \right)\right],\\
    \mathcal{L}_{Q_l} &= \mathbb{E}_{(s,a,s') \sim \mathcal{D}, g_s \sim p^\mathcal{D}_{\gamma_l}(\cdot \mid s, a))} \left[\left(Q_l(s,g_s,a) - r(s,g_s) - \gamma_l V_l(s',g_s)\right)^2\right], 
    \notag
    \end{aligned}
    \notag
\end{equation}
where $\tilde{Q}_l$ and $\tilde{V}_h$ denote the target networks, $\gamma_l = 1 - \frac{1}{n}$ denotes the low-level discount factor, and $\ell^2_\tau$ denotes the expectile loss $\ell^2_\tau(x) = |\tau - (x < 0)|x^2$. 

As for HIQL1vr, a high-level Q function is then learned only to allow DDPGBC high-level policy extraction:
\begin{equation}
    \mathcal{L}_{Q_h} = \mathbb{E}_{(s,\cdots, g_s) \sim \mathcal{D}, g \sim p^\mathcal{D}(\cdot \mid s, a))} \left[\left(Q_h(s,g, g_s) - V_h(g_s,g)\right)^2 \right].
    \notag
\end{equation}

Policies are then extracted using one of the following two loss functions for the low-level policy:
\begin{equation}
    \begin{aligned}
    \mathcal{L}^{\text{AWR}}_{\pi_l} &= -\mathbb{E}_{(s,a,s' \cdots g_s) \sim \mathcal{D}} \left[ e^{\alpha_l(Q_l(s, g_s, a)) - V_l(s, g_s))} \log \pi_l\left(a \mid s, g_s \right)\right],\\
    \mathcal{L}^{\text{DDPGBC}}_{\pi_l} &= -\mathbb{E}_{(s,a, s', \cdots g_s) \sim \mathcal{D}}\left[Q_l(s, g_s, \mu^{\pi_l}(s,g_s)) + \alpha_l \log \pi_l(a \mid s,g_s) \right], 
    \end{aligned}
    \notag
\end{equation}
and one of the following two loss functions for the high-level policy: 
\begin{equation}
    \begin{aligned}
    \mathcal{L}^{\text{AWR}}_{\pi_h} &= -\mathbb{E}_{(s \cdots g_s) \sim \mathcal{D}, g \sim p^\mathcal{D}_{\gamma}(\cdot \mid s)} \left[ e^{\alpha_h(Q_h(s, g, g_s)) - V_h(s, g))} \log \pi_h\left(g_s \mid s, g \right)\right],\\
    \mathcal{L}^{\text{DDPGBC}}_{\pi_h} &= -\mathbb{E}_{(s \cdots g_s) \sim \mathcal{D}, g \sim p^\mathcal{D}_{\gamma}(\cdot \mid s)}\left[Q_h(s, g, \mu^{\pi_h}(s,g)) + \alpha_h \log \pi_h(g_s \mid s,g) \right].
    \end{aligned}
    \notag
\end{equation}

\subsection{Hierarchical Implicit Q-Learning 2 Value Functions with Representation Learning (HIQL2vr)}
\label{app: hiql2vr}
Since HIQL1vr uses representation learning for the options, HIQL2vr learns option representations with the low-level value function. As in HIQL1vr (Section \ref{app: hiql1vr}), the representation is length-normalised. The high-level value function is trained as before, but the low-level value function, low-level Q function and high-level Q-function are trained using the following losses: 
\begin{equation}
    \begin{aligned}
    \mathcal{L}_{V_l, \phi_\omega} &= \mathbb{E}_{(s,a) \sim \mathcal{D}, g_s \sim p^\mathcal{D}_{\gamma_l}(\cdot \mid s, a))} \left[\ell^2_\tau\left(V_l(s, \hat\phi_\omega(s,g_s)) - \tilde{Q}_l(s,g_s,a) \right)\right],\\
    \mathcal{L}_{Q_l} &= \mathbb{E}_{(s,a,s') \sim \mathcal{D}, g_s \sim p^\mathcal{D}_{\gamma_l}(\cdot \mid s, a))} \left[\left(Q_l(s,g_s,a) - r(s,g_s) - \gamma_l V_l(s',\hat\phi_\omega(s',g_s))\right)^2\right], \\
    \mathcal{L}_{Q_h} &= \mathbb{E}_{(s,\cdots, g_s) \sim \mathcal{D}, g \sim p^\mathcal{D}(\cdot \mid s, a))} \left[\left(Q_h(s,g, \hat\phi_\omega(s,g_s)) - V_h(g_s,g)\right)^2 \right].
    \notag
    \end{aligned}
    \notag
\end{equation}
The policies are extracted using one of the following two loss functions for the low-level policy,
\begin{equation}
    \begin{aligned}
    \mathcal{L}^{\text{AWR}}_{\pi_l} &= -\mathbb{E}_{(s,a,s' \cdots g_s) \sim \mathcal{D}} \left[ e^{\alpha_l(Q_l(s, g_s, a)) - V_l(s, \hat\phi_\omega(s,g_s)))} \log \pi_l\left(a \mid s, \hat\phi_\omega(s,g_s) \right)\right],\\
    \mathcal{L}^{\text{DDPGBC}}_{\pi_l} &= -\mathbb{E}_{(s,a, s', \cdots g_s) \sim \mathcal{D}}\left[Q_l(s, g_s, \mu^{\pi_l}(s,\hat\phi_\omega(s,g_s))) + \alpha_l \log \pi_l(a \mid s, \hat\phi_\omega(s,g_s)) \right], 
    \end{aligned}
    \notag
\end{equation}
and one of the following two loss functions for the high-level policy: 
\begin{equation}
    \begin{aligned}
    \mathcal{L}^{\text{AWR}}_{\pi_h} &= -\mathbb{E}_{(s \cdots g_s) \sim \mathcal{D}, g \sim p^\mathcal{D}_{\gamma}(\cdot \mid s)} \left[ e^{\alpha_h(Q_h(s, g, \hat\phi_\omega(s,g_s))) - V_h(s, g))} \log \pi_h\left(\hat\phi_\omega(s,g_s) \mid s, g \right)\right],\\
    \mathcal{L}^{\text{DDPGBC}}_{\pi_h} &= -\mathbb{E}_{(s \cdots g_s) \sim \mathcal{D}, g \sim p^\mathcal{D}_{\gamma}(\cdot \mid s)}\left[Q_h(s, g, \mu^{\pi_h}(s,g)) + \alpha_h \log \pi_h(\hat\phi_\omega(s,g_s) \mid s,g) \right].
    \end{aligned}
    \notag
\end{equation} 

\subsection{Abstractive Reinforcement Learning Implicitly Learning Relativised Options (ARLi)}
\label{app: arli}
As presented in the main body of the paper, this is a minimal amendment to HIQL2vr, but, to learn relativised options via action similarity, option representations are now learned with the low-level policy. Equations are identical to HIQL2vr, except for the following low-level value functions:
\begin{equation}
    \begin{aligned}
    \mathcal{L}_{V_l} &= \mathbb{E}_{(s,a) \sim \mathcal{D}, g_s \sim p^\mathcal{D}_{\gamma_l}(\cdot \mid s, a))} \left[\ell^2_\tau\left(V_l(s, g_s)) - \tilde{Q}_l(s,g_s,a) \right)\right],\\
    \mathcal{L}_{Q_l} &= \mathbb{E}_{(s,a,s') \sim \mathcal{D}, g_s \sim p^\mathcal{D}_{\gamma_l}(\cdot \mid s, a))} \left[\left(Q_l(s,g_s,a) - r(s,g_s) - \gamma_l V_l(s',g_s)\right)^2\right],\\
    \notag
    \end{aligned}
    \notag
\end{equation}
and low-level policy functions:
\begin{equation}
    \begin{aligned}
    \mathcal{L}^{\text{AWR}}_{\pi_l, \phi_\omega} &= -\mathbb{E}_{(s,a,s' \cdots g_s) \sim \mathcal{D}} \left[ e^{\alpha_l(Q_l(s, g_s, a)) - V_l(s, g_s))} \log \pi_l\left(a \mid s, \hat\phi_\omega(s,g_s) \right)\right],\\
    \mathcal{L}^{\text{DDPGBC}}_{\pi_l, \phi_\omega} &= -\mathbb{E}_{(s,a, s', \cdots g_s) \sim \mathcal{D}}\left[Q_l(s, g_s, \mu^{\pi_l}(s,\hat\phi_\omega(s,g_s))) + \alpha_l \log \pi_l(a \mid s, \hat\phi_\omega(s,g_s)) \right].
    \end{aligned}
    \notag
\end{equation}

\begin{algorithm}[H]
\caption{Abstractive RL implicitly learning relativised options (ARLi)}
\label{alg: arli}
\small
\begin{algorithmic}
\State \textbf{Training}
\State Initialise low-level policy $\pi_l(a \mid s, \omega)$, high-level policy $\pi_h(\omega \mid s, g)$, and representation $\phi_\omega$.
\While{not converged}
    \State Sample batch $\mathcal{D}$
    \State \textcolor{clearsky}{$\triangleright$ \textbf{Hierarchical Policy}}
    \Statex \hspace{1em} $\left|\begin{aligned}
        &\hspace{1em} \text{Update low-level policy } \pi_l \text{ and representation } \phi_\omega \text{using } \pi_l(a \mid s, \hat{\phi}_\omega(s, g_s)) \text{ (Equation ~\ref{eq: low_awr})} \\
        &\hspace{1em} \omega \leftarrow \text{stopgrad}(\hat{\phi}_\omega(s, g_s)) \\
        &\hspace{1em} \text{Update high-level policy } \pi_h(\omega \mid s, g) \text{ (Equation ~\ref{eq: high_awr})}
    \end{aligned}\right.$
    \vspace{0.4em}
    \State \textcolor{clearsky}{$\triangleright$ \textbf{Hierarchical Value}}
    \Statex \hspace{1em} $\left|\begin{aligned}
        &\hspace{1em} \text{Update low-level value function } V_l(s, g_s) \text{ (Equation ~\ref{eq: low_v})} \\
        &\hspace{1em} \text{Update high-level value function } V_h(s, g) \text{ (Equation~\ref{eq: high_v})} \\
        &\hspace{1em} \text{Update critic } Q_l(s, g_s, a) \text{ (Equation ~\ref{eq: low_q})}
    \end{aligned}\right.$
    \vspace{0.4em}
\EndWhile
\State \Return $\pi_l, \pi_h, \phi_\omega$
\vspace{0.4em}
\State \textbf{Deployment (state $s$, goal $g$)}
\State Sample option from high-level policy $\omega \sim \pi_h(\cdot \mid s, g)$
\State Length-Normalise $\omega \leftarrow \frac{\omega}{\lVert \omega \rVert} \cdot \sqrt{d}$
\State Sample action from low-level policy $a \sim \pi_l(\cdot \mid s, \omega)$
\end{algorithmic}
\end{algorithm}

\subsection{Abstractive Reinforcement Learning Explicitly Enforcing Translational Invariance (ARLe)}
\label{app: arle}
ARL uses relativised options and relativised states for the low-level value function. Unlike ARLi, the low-level value and low-level critic now use \emph{relativised} goals, and options are explicitly relativised. As for ARLi, the option representations are learned with the low-level policy. The high-level value function is learned identically to HIQL2v and ARLi, but the low-level value function, low-level Q function and high-level Q function are learned as follows:
\begin{equation}
    \begin{aligned}
    \mathcal{L}_{V_l} &= \mathbb{E}_{(s,a) \sim \mathcal{D}, g_s \sim p^\mathcal{D}_{\gamma_l}(\cdot \mid s, a))} \left[\ell^2_\tau\left(V_l(g_s-s) - \tilde{Q}_l(s, g_s-s,a) \right)\right],\\
    \mathcal{L}_{Q_l} &= \mathbb{E}_{(s,a,s') \sim \mathcal{D}, g_s \sim p^\mathcal{D}_{\gamma_l}(\cdot \mid s, a))} \left[\left(Q_l(s,g_s-s,a) - r(s,g_s) - \gamma_l V_l(g_s-s')\right)^2\right],\\
    \mathcal{L}_{Q_h} &= \mathbb{E}_{(s,\cdots, g_s) \sim \mathcal{D}, g \sim p^\mathcal{D}(\cdot \mid s, a))} \left[\left(Q_h(s,g, \hat\phi_\omega(g_s-s)) - V_h(g_s,g)\right)^2 \right].
    \notag
    \end{aligned}
    \notag
\end{equation}
where, as before, $\tilde{Q}_l$ and $\tilde{V}_h$ denote the target networks, $\gamma_l = 1 - \frac{1}{n}$ denotes the low-level discount factor, and $\ell^2_\tau$ denotes the expectile loss $\ell^2_\tau(x) = |\tau - (x < 0)|x^2$. 
The policies are extracted using one of the following two loss functions for the low-level policy,
\begin{equation}
    \begin{aligned}
    \mathcal{L}^{\text{AWR}}_{\pi_l, \phi_\omega} &= -\mathbb{E}_{(s,a,s' \cdots g_s) \sim \mathcal{D}} \left[ e^{\alpha_l(Q_l(s, g_s-s, a)) - V_l(g_s-s))} \log \pi_l\left(a \mid s, \hat\phi_\omega(s, g_s) \right)\right],\\
    \mathcal{L}^{\text{DDPGBC}}_{\pi_l, \phi_\omega} &= -\mathbb{E}_{(s,a, s', \cdots g_s) \sim \mathcal{D}}\left[Q_l(s, g_s, \mu^{\pi_l}(s,\hat\phi_\omega(s, g_s))) + \alpha_l \log \pi_l(a \mid s, \hat\phi_\omega(s, g_s)) \right], 
    \end{aligned}
    \notag
\end{equation}
and one of the following two loss functions for the high-level policy: 
\begin{equation}
    \begin{aligned}
    \mathcal{L}^{\text{AWR}}_{\pi_h} &= -\mathbb{E}_{(s \cdots g_s) \sim \mathcal{D}, g \sim p^\mathcal{D}_{\gamma}(\cdot \mid s)} \left[ e^{\alpha_h(Q_h(s, g, \hat\phi_\omega(g_s-s))) - V_h(s, g))} \log \pi_h\left(\hat\phi_\omega(s, g_s) \mid s, g \right)\right],\\
    \mathcal{L}^{\text{DDPGBC}}_{\pi_h} &= -\mathbb{E}_{(s \cdots g_s) \sim \mathcal{D}, g \sim p^\mathcal{D}_{\gamma}(\cdot \mid s)}\left[Q_h(s, g, \mu^{\pi_h}(s,g)) + \alpha_h \log \pi_g(\hat\phi_\omega(s, g_s) \mid s,g) \right].
    \end{aligned}
    \notag
\end{equation} 

\begin{algorithm}[t]
\caption{Abstractive RL explicitly enforcing translation invariance (ARLe)}
\label{alg: arle}
\small
\begin{algorithmic}
\State \textbf{Training}
\State Initialise low-level policy $\pi_l(a \mid s, \omega)$, high-level policy $\pi_h(\omega \mid s, g)$, and representation $\phi_\omega$.
\While{not converged}
    \State Sample batch from $\mathcal{D}$
    \State \textcolor{clearsky}{$\triangleright$ \textbf{Hierarchical Policy}}
    \Statex \hspace{1em} $\left|\begin{aligned}
        &\hspace{1em} \text{Update low-level policy } \pi_l \text{ and representation } \phi_\omega \text{using } \pi_l(a \mid s, \hat{\phi}_\omega(s, g_s)) \text{ (Equation ~\ref{eq: low_awr})} \\
        &\hspace{1em}\omega \leftarrow \text{stopgrad}(\hat{\phi}_\omega(s, g_s)) \\
        &\hspace{1em} \text{Update high-level policy } \pi_h(\omega \mid s, g) \text{ (Equation ~\ref{eq: high_awr})}
    \end{aligned}\right.$
    \vspace{0.4em}
    \State \textcolor{clearsky}{$\triangleright$ \textbf{Hierarchical Value}}
    \Statex \hspace{1em} $\left|\begin{aligned}
        &\hspace{1em} \text{Update low-level value function } V_l(g_s - s) \text{ (Equation ~\ref{eq: low_v})} \\
        &\hspace{1em} \text{Update high-level value function } V_h(s, g) \text{ (Equation~\ref{eq: high_v})} \\
        &\hspace{1em} \text{Update critic } Q_l(s, g_s - s, a) \text{ (Equation ~\ref{eq: low_q})}
    \end{aligned}\right.$
    \vspace{0.4em}
\EndWhile
\State \Return $\pi_l, \pi_h, \phi_\omega$
\vspace{0.4em}
\State \textbf{Deployment (state $s$, goal $g$)}
\State Sample option from high-level policy $\omega \sim \pi_h(\cdot \mid s, g)$
\State Soft-Normalise $\omega \leftarrow \frac{\omega \cdot \tanh(\lVert \omega \rVert)}{\lVert \omega \rVert} \cdot \sqrt{d} $
\State Sample action from low-level policy $a \sim \pi_l(\cdot \mid s, \omega)$
\end{algorithmic}
\end{algorithm}

\newpage
\section{Full Results}
\label{app: results}
\begin{table}[H]
\centering
\caption{\textbf{Full Results 1.} We report each method’s average (binary) success rate (\%) across the five test-time goals on each task. Bootstrapped 95\% CI over 4 seeds and 20 evaluation runs. Blue bold indicates the highest mean; black bold overlapping confidence intervals.}
\footnotesize
\begin{tabular}{ll cccccc}
\toprule
\textbf{Environment} & \textbf{Task} & \textbf{IQL} & \textbf{HIQL1vr} & \textbf{HIQL2v} & \textbf{HIQL2vr} & \textbf{ARLi} & \textbf{ARLe} \\
\midrule
\multirow{6}{*}{pointmaze-giant-navigate-v0} & 1 & 0$\pm$0 & 1$\pm$2 & 6$\pm$9 & 2$\pm$4 & 38$\pm$38 & 4$\pm$4 \\
& 2 & 0$\pm$0 & 66$\pm$22 & 64$\pm$24 & 75$\pm$19 & 86$\pm$7 & 31$\pm$14 \\
& 3 & 0$\pm$0 & 49$\pm$21 & 1$\pm$2 & 24$\pm$16 & 8$\pm$8 & 2$\pm$4 \\
& 4 & 0$\pm$0 & 71$\pm$33 & 4$\pm$6 & 60$\pm$20 & 10$\pm$10 & 22$\pm$21 \\
& 5 & 0$\pm$0 & 89$\pm$7 & 31$\pm$19 & 69$\pm$18 & 96$\pm$4 & 21$\pm$14 \\
& Overall & 0$\pm$0 & \textcolor{clearsky!95!black}{\textbf{55$\pm$11}} & 21$\pm$7 & \textbf{46$\pm$6} & \textbf{48$\pm$8} & 16$\pm$2 \\
\cmidrule(lr){2-8}
\multirow{6}{*}{pointmaze-giant-stitch-v0} & 1 & 0$\pm$0 & 0$\pm$0 & 0$\pm$0 & 0$\pm$0 & 0$\pm$0 & 0$\pm$0 \\
& 2 & 0$\pm$0 & 0$\pm$0 & 0$\pm$0 & 0$\pm$0 & 0$\pm$0 & 0$\pm$0 \\
& 3 & 0$\pm$0 & 0$\pm$0 & 0$\pm$0 & 0$\pm$0 & 0$\pm$0 & 0$\pm$0 \\
& 4 & 0$\pm$0 & 0$\pm$0 & 0$\pm$0 & 0$\pm$0 & 0$\pm$0 & 0$\pm$0 \\
& 5 & 0$\pm$0 & 0$\pm$0 & 0$\pm$0 & 0$\pm$0 & 0$\pm$0 & 0$\pm$0 \\
& Overall & 0$\pm$0 & 0$\pm$0 & 0$\pm$0 & 0$\pm$0 & 0$\pm$0 & 0$\pm$0 \\
\midrule
\multirow{6}{*}{antmaze-giant-navigate-v0} & 1 & 0$\pm$0 & 18$\pm$8 & 19$\pm$11 & 29$\pm$9 & 18$\pm$5 & 25$\pm$4 \\
& 2 & 1$\pm$2 & 44$\pm$19 & 56$\pm$6 & 50$\pm$10 & 48$\pm$8 & 42$\pm$8 \\
& 3 & 0$\pm$0 & 34$\pm$7 & 25$\pm$8 & 39$\pm$11 & 51$\pm$18 & 35$\pm$10 \\
& 4 & 0$\pm$0 & 30$\pm$6 & 44$\pm$7 & 54$\pm$14 & 56$\pm$9 & 44$\pm$6 \\
& 5 & 0$\pm$0 & 64$\pm$6 & 57$\pm$9 & 71$\pm$16 & 66$\pm$6 & 64$\pm$11 \\
& Overall & 0$\pm$0 & 38$\pm$3 & \textbf{40$\pm$3} & \textcolor{clearsky!95!black}{\textbf{48$\pm$6}} & \textcolor{clearsky!95!black}{\textbf{48$\pm$3}} & \textbf{42$\pm$4} \\
\cmidrule(lr){2-8}
\multirow{6}{*}{antmaze-giant-stitch-v0} & 1 & 0$\pm$0 & 8$\pm$8 & 9$\pm$4 & 30$\pm$5 & 44$\pm$14 & 26$\pm$7 \\
& 2 & 0$\pm$0 & 6$\pm$4 & 30$\pm$12 & 26$\pm$11 & 22$\pm$5 & 14$\pm$4 \\
& 3 & 0$\pm$0 & 2$\pm$2 & 5$\pm$8 & 4$\pm$4 & 15$\pm$6 & 10$\pm$13 \\
& 4 & 0$\pm$0 & 16$\pm$16 & 16$\pm$11 & 36$\pm$9 & 54$\pm$14 & 45$\pm$11 \\
& 5 & 0$\pm$0 & 4$\pm$4 & 15$\pm$6 & 6$\pm$2 & 26$\pm$21 & 10$\pm$0 \\
& Overall & 0$\pm$0 & 7$\pm$7 & 15$\pm$3 & 20$\pm$3 & \textcolor{clearsky!95!black}{\textbf{32$\pm$7}} & 21$\pm$1 \\
\cmidrule(lr){2-8}
\multirow{6}{*}{antmaze-teleport-stitch-v0} & 1 & 38$\pm$8 & 41$\pm$7 & 55$\pm$8 & 45$\pm$9 & 42$\pm$9 & 45$\pm$4 \\
& 2 & 55$\pm$4 & 36$\pm$11 & 61$\pm$6 & 46$\pm$7 & 45$\pm$13 & 55$\pm$11 \\
& 3 & 55$\pm$9 & 16$\pm$4 & 31$\pm$12 & 34$\pm$4 & 35$\pm$18 & 29$\pm$4 \\
& 4 & 45$\pm$9 & 24$\pm$9 & 22$\pm$7 & 39$\pm$11 & 27$\pm$4 & 36$\pm$14 \\
& 5 & 51$\pm$6 & 26$\pm$9 & 46$\pm$11 & 34$\pm$11 & 31$\pm$6 & 41$\pm$6 \\
& Overall & \textcolor{clearsky!95!black}{\textbf{49$\pm$2}} & 29$\pm$4 & 43$\pm$3 & 40$\pm$6 & 36$\pm$5 & 41$\pm$4 \\
\midrule
\multirow{6}{*}{humanoidmaze-giant-navigate-v0} & 1 & 1$\pm$2 & 16$\pm$9 & 8$\pm$8 & 2$\pm$4 & 40$\pm$9 & 29$\pm$11 \\
& 2 & 2$\pm$2 & 36$\pm$19 & 29$\pm$14 & 16$\pm$14 & 50$\pm$6 & 46$\pm$4 \\
& 3 & 0$\pm$0 & 12$\pm$5 & 11$\pm$6 & 10$\pm$9 & 41$\pm$9 & 32$\pm$5 \\
& 4 & 0$\pm$0 & 19$\pm$18 & 11$\pm$6 & 15$\pm$12 & 57$\pm$10 & 40$\pm$10 \\
& 5 & 0$\pm$0 & 26$\pm$9 & 1$\pm$2 & 10$\pm$13 & 56$\pm$14 & 45$\pm$4 \\
& Overall & 1$\pm$1 & 22$\pm$11 & 12$\pm$5 & 11$\pm$10 & \textcolor{clearsky!95!black}{\textbf{49$\pm$6}} & 38$\pm$4 \\
\cmidrule(lr){2-8}
\multirow{6}{*}{humanoidmaze-giant-stitch-v0} & 1 & 0$\pm$0 & 5$\pm$8 & 0$\pm$0 & 0$\pm$0 & 12$\pm$5 & 11$\pm$6 \\
& 2 & 2$\pm$2 & 11$\pm$8 & 2$\pm$2 & 0$\pm$0 & 31$\pm$16 & 21$\pm$9 \\
& 3 & 0$\pm$0 & 2$\pm$2 & 0$\pm$0 & 0$\pm$0 & 11$\pm$2 & 8$\pm$4 \\
& 4 & 0$\pm$0 & 0$\pm$0 & 0$\pm$0 & 0$\pm$0 & 9$\pm$9 & 9$\pm$7 \\
& 5 & 0$\pm$0 & 0$\pm$0 & 0$\pm$0 & 0$\pm$0 & 2$\pm$2 & 4$\pm$2 \\
& Overall & 0$\pm$0 & 4$\pm$3 & 0$\pm$0 & 0$\pm$0 & \textcolor{clearsky!95!black}{\textbf{13$\pm$2}} & \textbf{10$\pm$3} \\
\bottomrule
\end{tabular}
\end{table}

\begin{table}[H]
\centering
\caption{\textbf{Full Results 2.} We report each method’s average (binary) success rate (\%) across the five test-time goals on each task. Bootstrapped 95\% CI over 4 seeds and 20 evaluation runs. Blue bold indicates the highest mean; black bold overlapping confidence intervals.}
\footnotesize
\begin{tabular}{ll cccccc}
\toprule
\textbf{Environment} & \textbf{Task} & \textbf{IQL} & \textbf{HIQL1vr} & \textbf{HIQL2v} & \textbf{HIQL2vr} & \textbf{ARLi} & \textbf{ARLe} \\
\midrule
\multirow{6}{*}{cube-double-play-v0} & 1 & 94$\pm$6 & 8$\pm$2 & 0$\pm$0 & 14$\pm$2 & 96$\pm$4 & 94$\pm$4 \\
& 2 & 46$\pm$11 & 0$\pm$0 & 0$\pm$0 & 0$\pm$0 & 59$\pm$6 & 78$\pm$11 \\
& 3 & 59$\pm$8 & 0$\pm$0 & 0$\pm$0 & 0$\pm$0 & 52$\pm$5 & 68$\pm$4 \\
& 4 & 15$\pm$6 & 0$\pm$0 & 0$\pm$0 & 0$\pm$0 & 14$\pm$7 & 35$\pm$6 \\
& 5 & 34$\pm$9 & 0$\pm$0 & 0$\pm$0 & 0$\pm$0 & 44$\pm$6 & 61$\pm$9 \\
& Overall & 50$\pm$3 & 2$\pm$0 & 0$\pm$0 & 3$\pm$0 & 53$\pm$2 & \textcolor{clearsky!95!black}{\textbf{67$\pm$3}} \\
\cmidrule(lr){2-8}
\multirow{6}{*}{cube-triple-play-v0} & 1 & 51$\pm$7 & 35$\pm$16 & 0$\pm$0 & 4$\pm$4 & 70$\pm$10 & 69$\pm$9 \\
& 2 & 1$\pm$2 & 0$\pm$0 & 0$\pm$0 & 0$\pm$0 & 1$\pm$2 & 2$\pm$2 \\
& 3 & 4$\pm$2 & 0$\pm$0 & 0$\pm$0 & 0$\pm$0 & 0$\pm$0 & 2$\pm$4 \\
& 4 & 0$\pm$0 & 0$\pm$0 & 0$\pm$0 & 0$\pm$0 & 0$\pm$0 & 0$\pm$0 \\
& 5 & 0$\pm$0 & 0$\pm$0 & 0$\pm$0 & 0$\pm$0 & 0$\pm$0 & 0$\pm$0 \\
& Overall & 11$\pm$2 & 7$\pm$3 & 0$\pm$0 & 1$\pm$1 & \textbf{14$\pm$2} & \textcolor{clearsky!95!black}{\textbf{15$\pm$3}} \\
\cmidrule(lr){2-8}
\multirow{6}{*}{cube-quadruple-play-v0} & 1 & 0$\pm$0 & 0$\pm$0 & 0$\pm$0 & 0$\pm$0 & 4$\pm$2 & 2$\pm$2 \\
& 2 & 0$\pm$0 & 0$\pm$0 & 0$\pm$0 & 0$\pm$0 & 0$\pm$0 & 0$\pm$0 \\
& 3 & 0$\pm$0 & 0$\pm$0 & 0$\pm$0 & 0$\pm$0 & 1$\pm$2 & 0$\pm$0 \\
& 4 & 0$\pm$0 & 0$\pm$0 & 0$\pm$0 & 0$\pm$0 & 0$\pm$0 & 0$\pm$0 \\
& 5 & 0$\pm$0 & 0$\pm$0 & 0$\pm$0 & 0$\pm$0 & 0$\pm$0 & 0$\pm$0 \\
& Overall & 0$\pm$0 & 0$\pm$0 & 0$\pm$0 & 0$\pm$0 & \textcolor{clearsky!95!black}{\textbf{1$\pm$0}} & 0$\pm$0 \\
\midrule
\multirow{6}{*}{puzzle-3x3-play-v0} & 1 & 100$\pm$0 & 100$\pm$0 & 0$\pm$0 & 100$\pm$0 & 100$\pm$0 & 100$\pm$0 \\
& 2 & 100$\pm$0 & 14$\pm$21 & 0$\pm$0 & 15$\pm$12 & 88$\pm$12 & 45$\pm$34 \\
& 3 & 99$\pm$2 & 4$\pm$6 & 0$\pm$0 & 0$\pm$0 & 78$\pm$19 & 22$\pm$30 \\
& 4 & 100$\pm$0 & 2$\pm$4 & 0$\pm$0 & 2$\pm$2 & 77$\pm$28 & 21$\pm$28 \\
& 5 & 100$\pm$0 & 15$\pm$12 & 0$\pm$0 & 1$\pm$2 & 89$\pm$11 & 29$\pm$32 \\
& Overall & \textcolor{clearsky!95!black}{\textbf{100$\pm$0}} & 27$\pm$8 & 0$\pm$0 & 24$\pm$3 & 86$\pm$14 & 44$\pm$25 \\
\cmidrule(lr){2-8}
\multirow{6}{*}{puzzle-4x4-play-v0} & 1 & 50$\pm$10 & 66$\pm$36 & 0$\pm$0 & 29$\pm$12 & 91$\pm$9 & 100$\pm$0 \\
& 2 & 6$\pm$4 & 32$\pm$24 & 0$\pm$0 & 15$\pm$6 & 70$\pm$15 & 75$\pm$18 \\
& 3 & 38$\pm$3 & 59$\pm$32 & 0$\pm$0 & 18$\pm$8 & 81$\pm$7 & 90$\pm$9 \\
& 4 & 29$\pm$13 & 48$\pm$29 & 0$\pm$0 & 10$\pm$4 & 72$\pm$18 & 89$\pm$6 \\
& 5 & 29$\pm$7 & 41$\pm$25 & 0$\pm$0 & 12$\pm$5 & 72$\pm$13 & 85$\pm$5 \\
& Overall & 30$\pm$3 & 49$\pm$27 & 0$\pm$0 & 17$\pm$6 & \textbf{78$\pm$11} & \textcolor{clearsky!95!black}{\textbf{88$\pm$6}} \\
\cmidrule(lr){2-8}
\multirow{6}{*}{puzzle-4x5-play-v0} & 1 & 72$\pm$12 & 82$\pm$11 & 0$\pm$0 & 62$\pm$15 & 90$\pm$13 & 68$\pm$36 \\
& 2 & 1$\pm$2 & 2$\pm$4 & 0$\pm$0 & 0$\pm$0 & 0$\pm$0 & 0$\pm$0 \\
& 3 & 0$\pm$0 & 0$\pm$0 & 0$\pm$0 & 0$\pm$0 & 0$\pm$0 & 0$\pm$0 \\
& 4 & 0$\pm$0 & 0$\pm$0 & 0$\pm$0 & 0$\pm$0 & 0$\pm$0 & 0$\pm$0 \\
& 5 & 0$\pm$0 & 0$\pm$0 & 0$\pm$0 & 0$\pm$0 & 0$\pm$0 & 0$\pm$0 \\
& Overall & \textbf{15$\pm$3} & \textbf{17$\pm$3} & 0$\pm$0 & \textbf{12$\pm$3} & \textcolor{clearsky!95!black}{\textbf{18$\pm$3}} & \textbf{14$\pm$7} \\
\cmidrule(lr){2-8}
\multirow{6}{*}{puzzle-4x6-play-v0} & 1 & 51$\pm$9 & 0$\pm$0 & 0$\pm$0 & 76$\pm$6 & 65$\pm$32 & 45$\pm$45 \\
& 2 & 15$\pm$6 & 0$\pm$0 & 0$\pm$0 & 16$\pm$9 & 2$\pm$2 & 0$\pm$0 \\
& 3 & 0$\pm$0 & 0$\pm$0 & 0$\pm$0 & 0$\pm$0 & 0$\pm$0 & 0$\pm$0 \\
& 4 & 0$\pm$0 & 0$\pm$0 & 0$\pm$0 & 0$\pm$0 & 0$\pm$0 & 0$\pm$0 \\
& 5 & 0$\pm$0 & 0$\pm$0 & 0$\pm$0 & 0$\pm$0 & 0$\pm$0 & 0$\pm$0 \\
& Overall & 13$\pm$1 & 0$\pm$0 & 0$\pm$0 & \textcolor{clearsky!95!black}{\textbf{18$\pm$2}} & \textbf{14$\pm$7} & \textbf{9$\pm$9} \\
\midrule
\multirow{6}{*}{scene-play-v0} & 1 & 31$\pm$6 & 22$\pm$5 & 26$\pm$23 & 26$\pm$7 & 50$\pm$6 & 44$\pm$13 \\
& 2 & 16$\pm$11 & 5$\pm$4 & 11$\pm$4 & 6$\pm$4 & 16$\pm$11 & 18$\pm$3 \\
& 3 & 6$\pm$8 & 8$\pm$8 & 5$\pm$8 & 10$\pm$4 & 19$\pm$4 & 12$\pm$2 \\
& 4 & 9$\pm$6 & 16$\pm$6 & 11$\pm$4 & 19$\pm$9 & 30$\pm$8 & 26$\pm$9 \\
& 5 & 1$\pm$2 & 9$\pm$6 & 8$\pm$5 & 4$\pm$4 & 8$\pm$5 & 6$\pm$4 \\
& Overall & 13$\pm$2 & 12$\pm$3 & 12$\pm$7 & 13$\pm$1 & \textcolor{clearsky!95!black}{\textbf{24$\pm$3}} & \textbf{21$\pm$3} \\
\bottomrule
\end{tabular}
\end{table}

\begin{figure}[H]
    \centering
    \includegraphics[width=0.67\linewidth]{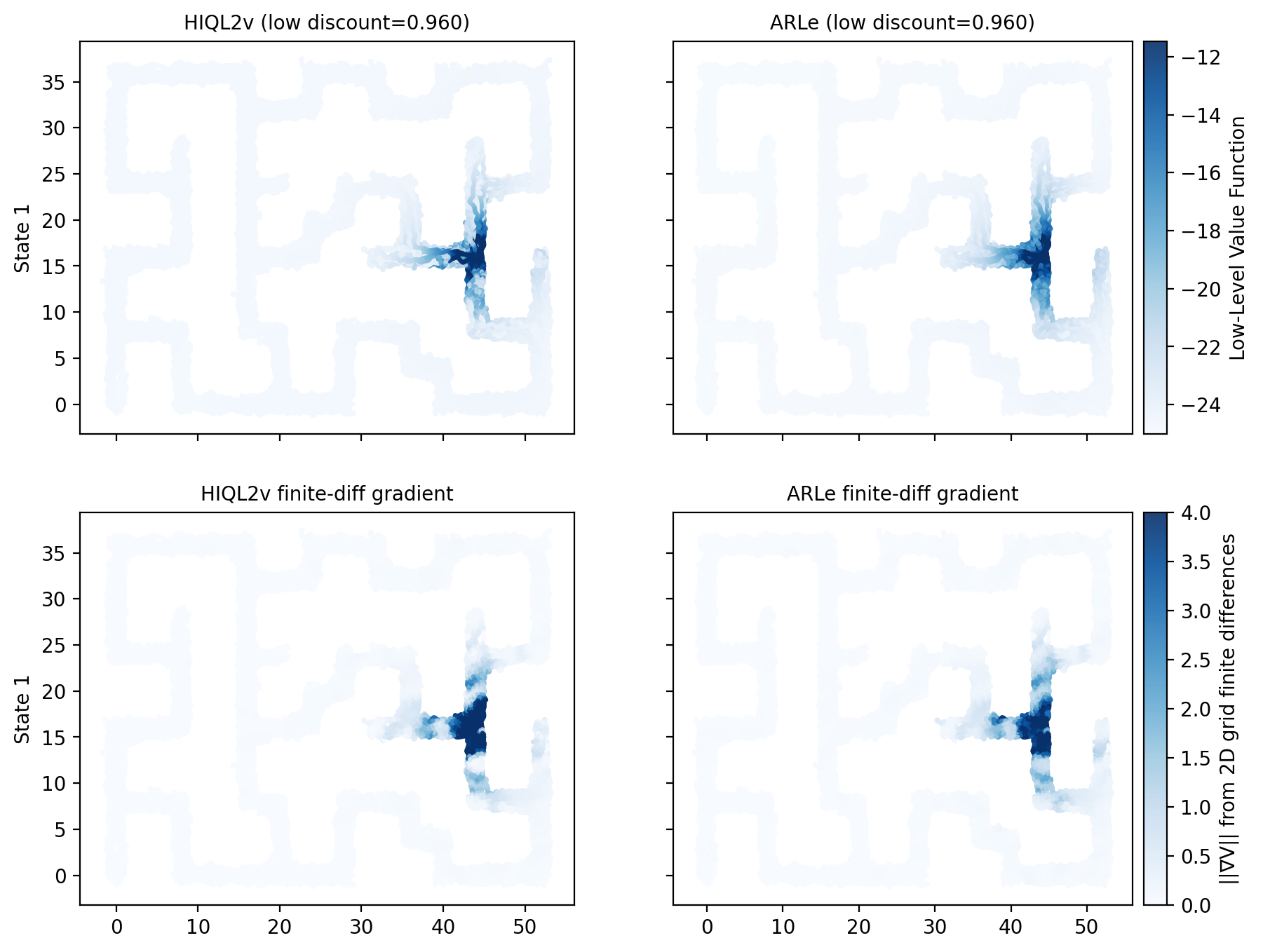}
    \caption{\textbf{Low-level} value functions (\textbf{top}) and \textbf{gradient} of low-level value function (\textbf{bottom}). IQL and HIQL1vr are excluded as they have a single value function.}
    \label{fig: low_level_value}
\end{figure}

\begin{figure}[H]
    \centering
    \includegraphics[width=1.0\linewidth]{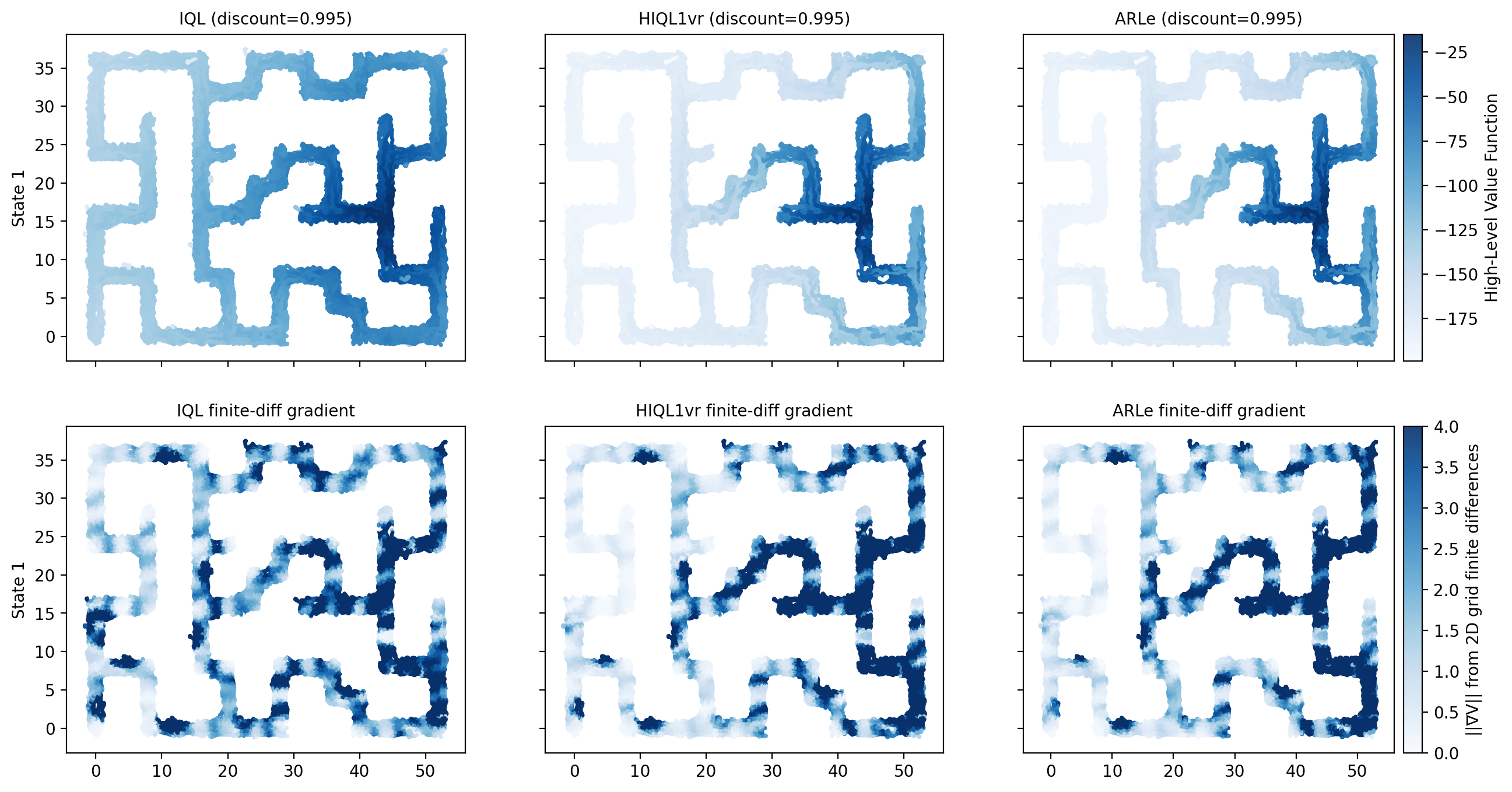}
    \caption{\textbf{High-level} value functions (\textbf{top}) and \textbf{gradient} of high-level value function (\textbf{bottom}). HIQL2v and HIQL2vr are excluded, as they have identical ones to ARL.}
    \label{fig: high_level_value}
\end{figure}

 \begin{figure}[H]
     \centering
     \includegraphics[width=1.0\linewidth]{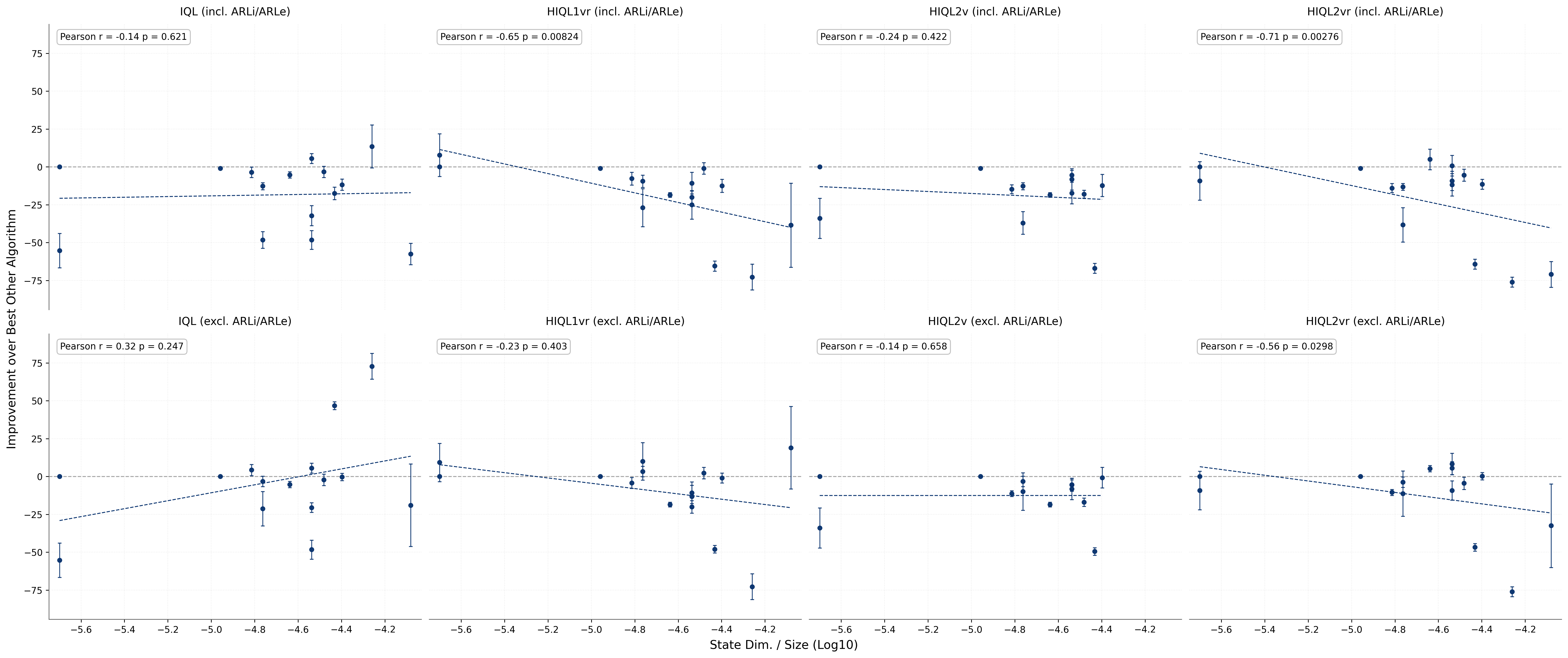}
     \caption{\textbf{Performance improvements} over next best performing algorithm against number of state dimensions per dataset sample: \textbf{IQL} (\textbf{left}), \textbf{HIQL1vr} (\textbf{centre left}), \textbf{HIQL2v} (\textbf{centre right}) and \textbf{HIQL2vr} (\textbf{right}). Including ARLi and ARLe (\textbf{top}), and excluding ARLi and ARLe (\textbf{bottom}). Bootstrapped 95\% CI over 4 seeds and 20 evaluation runs.}
     \label{fig: full_correlation}
 \end{figure}

\newpage
\section{Experimental Details}
\label{app: details}
We release all code and hyperparameters in the following repository: \url{https://github.com/CWibault/arl}. 

\paragraph{Datasets.} We use the standard OGBench datasets. States are randomly and uniformly sampled from the dataset. Goals for the value function learning and policy extraction are sampled using a certain probability of sampling the current state $p^\mathcal{D}_{\text{cur}}$, from the current trajectory $p^\mathcal{D}_{\text{traj}}$ (geometrically, according to the discount factor, or uniformally), or randomly from the dataset $p^\mathcal{D}_{\text{rand}}$. waypoints $g_s$ are taken as the states $n$ steps ahead of the current state.  

\begin{table}[H]
\centering
\caption{\textbf{Environment Characteristics.} Environment properties to provide intuition for interpreting results.}
\footnotesize
\begin{tabular}{lcccc}
\toprule
\textbf{Task} & \textbf{State Dim.} & \textbf{Action Dim.} & \textbf{Max. Episode Length} & \textbf{Dataset Size} \\
\midrule
pointmaze-giant & 2 & 2 & 1000 & 1M\\
\midrule
antmaze-giant & 29 & 8 & 1000 & 1M\\
antmaze-teleport & 29 & 8 & 1000 & 1M\\
\midrule
humanoidmaze-large & 69 & 21 & 1000 & 4M\\
humanoidmaze-giant & 69 & 21 & 4000 & 4M\\
\midrule
cube-double & 37 & 5 & 500 & 1M\\
cube-triple & 46 & 5 & 1000 & 3M\\
cube-quadruple & 55 & 5 & 1000 & 5M\\
\midrule
puzzle-3x3 & 55 & 5 & 500 & 1M\\
puzzle-4x4 & 83 & 5 & 500 & 1M\\
puzzle-4x5 & 99 & 5 & 1000 & 3M\\
puzzle-4x6 & 115 & 5 & 1000 & 5M\\
\midrule
scene-play & 40 & 5 & 750 & 1M\\
\bottomrule
\end{tabular}
\label{tab: env_properties}
\end{table}

\paragraph{Reward Relabelling.} Unlike prior work on OGBench \citep{park_ogbench_2025}, we use the original environment reward functions for relabeling rewards rather than using a binary indicator of the state-index to ensure that the agent remains focused on the primary task objectives and prevents the value function from becoming overly specific to irrelevant dimensions. To ensure fair comparison, this relabeling strategy is applied consistently across all algorithms and tasks. Note that assuming access to the environment reward function in robotic tasks is an entirely valid assumption.

\paragraph{Hyperparameters.} We provide the full list of hyperparameters. We follow those from \citet{park_ogbench_2025} and \citet{park_horizon_2025}. Notably, while these parameters were specifically tuned for HIQL, we apply them to ARL without further adjustment. The fact that ARL achieves strong performance using parameters optimised for a different algorithm demonstrates its robustness. We use DDPGBC with a behaviour cloning strength of 0.1 to extract the high-level policy in manipulation tasks, which allows for more extrapolation \citep{park_is_2024}. This was generally found to outperform AWR. 

\begin{table}[H]
\centering
\caption{Hyperparameters}
\footnotesize
\begin{tabular}{ll}
\hline
\textbf{Hyperparameter} & \textbf{Value} \\
\hline
Gradient steps & $10^6$ \\
Optimiser & Adam \citep{kingma2017adammethodstochasticoptimization} \\
Learning rate & 0.0003 \\
Batch size & 1024 \\
Layer Normalisation \citep{ba2016layernormalization} & True \\
Nonlinearity & GELU \citep{hendrycks2018gelu} \\
Value MLP & [1024, 1024, 1024, 1024] \\
Actor MLP & [1024, 1024, 1024, 1024] \\
Representation MLP & [512, 512, 512] \\
Representation Dimension & 10 \\
Target network update rate & 0.005 \\
IQL Expectile $\tau$ & 0.9 (IQL), 0.7 (HIQL1vr, HIQL2v, HIQL2vr, ARLi, ARLe) \\
Value ratio $(p^{\mathcal{D}}_\text{cur},p^{\mathcal{D}}_\text{traj}, p^{\mathcal{D}}_\text{rand},p^{\mathcal{D}}_\text{geom})$
& $(0.2, 0.5, 0.3, 0)$ \\
Low Value ratio $(p^{\mathcal{D}}_\text{cur},p^{\mathcal{D}}_\text{traj}, p^{\mathcal{D}}_\text{rand},p^{\mathcal{D}}_\text{geom})$
& $(0.10, 0.85, 0.05, 1)$ \\
High Value ratio $(p^{\mathcal{D}}_\text{cur},p^{\mathcal{D}}_\text{traj}, p^{\mathcal{D}}_\text{rand},p^{\mathcal{D}}_{\text{geom}})$
& $(0.2, 0.5, 0.3, 0)$ \\
Policy ratio $(p^{\mathcal{D}}_\text{cur},p^{\mathcal{D}}_\text{traj}, p^{\mathcal{D}}_\text{rand},p^{\mathcal{D}}_\text{geom})$ 
& $(0.0, 0.5, 0.5, 1)$\\
\hline
\end{tabular}
\end{table}

\begin{table}[H]
\centering
\caption{Task-Specific Hyperparameters}
\footnotesize
\begin{tabular}{ll ccccccc}
\hline
\textbf{Task} & $n$ & $\gamma$ & \textbf{Loss $\pi$} & $\alpha$ & \textbf{Loss $\pi_l$} & $\alpha_l$ & \textbf{Loss $\pi_h$} & $\alpha_h$ \\
\hline
pointmaze-giant-navigate-v0 & 25 & 0.995 & DDPGBC & 0.1 & AWR & 3.0 & AWR & 3.0 \\
pointmaze-giant-stitch-v0   & 25 & 0.995 & DDPGBC & 0.1 & AWR & 3.0 & AWR & 3.0 \\
\midrule
antmaze-giant-navigate-v0   & 25 & 0.995 & DDPGBC & 0.1 & AWR & 3.0 & AWR & 3.0 \\
antmaze-giant-stitch-v0     & 25 & 0.995 & DDPGBC & 0.1 & AWR & 3.0 & AWR & 3.0 \\
antmaze-teleport-stitch-v0  & 25 & 0.990 & DDPGBC & 0.1 & AWR & 3.0 & AWR & 3.0 \\
\midrule
humanoidmaze-giant-navigate-v0 & 100 & 0.999 & DDPGBC & 0.1 & AWR & 3.0 & AWR & 3.0 \\
humanoidmaze-giant-stitch-v0   & 100 & 0.999 & DDPGBC & 0.1 & AWR & 3.0 & AWR & 3.0 \\
\midrule
cube-double-play-v0      & 25 & 0.99 & DDPGBC & 1.0 & AWR & 3.0 & DDPGBC & 0.1 \\
cube-triple-play-v0      & 25 & 0.99 & DDPGBC & 1.0 & AWR & 3.0 & DDPGBC & 0.1 \\
cube-quadruple-play-v0   & 25 & 0.99 & DDPGBC & 1.0 & AWR & 3.0 & DDPGBC & 0.1 \\
\midrule
puzzle-3x3-play-v0 & 25 & 0.99 & DDPGBC & 1.0 & AWR & 3.0 & DDPGBC & 0.1 \\
puzzle-4x4-play-v0 & 25 & 0.99 & DDPGBC & 1.0 & AWR & 3.0 & DDPGBC & 0.1 \\
puzzle-4x5-play-v0 & 25 & 0.99 & DDPGBC & 1.0 & AWR & 3.0 & DDPGBC & 0.1 \\
puzzle-4x6-play-v0 & 25 & 0.99 & DDPGBC & 1.0 & AWR & 3.0 & DDPGBC & 0.1 \\
\midrule
scene-play-v0 & 25 & 0.99 & DDPGBC & 1.0 & AWR & 3.0 & DDPGBC & 0.1 \\
\hline
\end{tabular}
\end{table}

The low-level discount factor is computed as $\gamma_l = 1 - \frac{1}{n}$. The high-level discount factor is computed as $\gamma_h = \gamma^n$. 

\newpage
\section{Limitations}
\label{app: limitations}
ARLe’s performance depends on the alignment between its inductive bias (assuming translational invariance) and the environment's structure. We discuss these limitations with the experiments (Section \ref{sec: experiments}) and in the conclusion (Section \ref{sec: conclusion}).

Like other prior methods learning action-free value functions, learning an action-free high-level value function biases our instantiations of ARL towards being optimistic in stochastic environments \citep{park_hiql_2024}. Such optimism bias could be addressed by disentangling controllable parts of the state \citep{villaflor2022addressingoptimismbiassequence}, but we leave this to future work. We also note that, since only the high-level value function is action-free, performance degradation compared to IQL for both ARLe and ARLi in the stochastic environment (\textit{antmaze-teleport-stitch-v0}) is less significant than for HIQL1vr, for example.

Including more than 15 environments would have helped to strengthen our hypothesis in Section \ref{sec: conclusion}, but we were limited by compute resources. 

\section{Compute}
\label{app: compute}
All experiments were conducted on NVIDIA L40 GPUs, lasting 3 hours per run, including evaluation.

\section*{Impact Statement}
\label{app: impact}
This paper presents work whose goal is to advance the field of machine learning. There are many potential societal consequences of our work, none of which we feel must be specifically highlighted here.

\newpage
\end{document}